\documentclass[lettersize,journal]{IEEEtran}

\usepackage{pifont}% http://ctan.org/pkg/pifont
\usepackage{bbm}
\usepackage{makecell}
\usepackage{xcolor}
\usepackage[numbers]{natbib}
\usepackage[bookmarks=true]{hyperref}
\usepackage{array}
\usepackage[caption=false,font=normalsize,labelfont=sf,textfont=sf]{subfig}
\usepackage{textcomp}
\usepackage{stfloats}
\usepackage{verbatim}

\usepackage{algorithm}
\usepackage[noend]{algpseudocode}
\usepackage{tabularx}

\usepackage{amsmath,amssymb,amsfonts,balance,booktabs,color,enumerate,float,graphicx,hyperref,multicol,multirow,siunitx,times,subcaption,mathtools,colortbl,caption,url}

\usepackage[final]{changes}
\newcommand{\method}{RobotMover}

\def\BibTeX{{\rm B\kern-.05em{\sc i\kern-.025em b}\kern-.08em
    T\kern-.1667em\lower.7ex\hbox{E}\kern-.125emX}}

\begin{document}
\title{RobotMover: Learning to Move \\ Large Objects from Human Demonstrations}
% \author{Author Names Omitted for Anonymous Review.
\author{Tianyu Li$^{1,*}$, Joanne Truong$^{2}$, Jimmy Yang$^{2}$, Alexander Clegg$^{2}$, Akshara Rai$^{2}$, Sehoon Ha$^{1,+}$, Xavier Puig$^{2,+}$
\authorblockA{
    $^{1}$Georgia Institute of Technology, $^{2}$FAIR, Meta \\
    }
\thanks{
    $^{*}$Work done during an internship at FAIR, Meta. \\
    }
}

\markboth{Journal of \LaTeX\ Class Files,~Vol.~XX, No.~X, XXXXXXX~202X}%
{RobotMover: Learning to Move Large Objects from Human Demonstrations}

\makeatletter
\let\@oldmaketitle\@maketitle%
\renewcommand{\@maketitle}{\@oldmaketitle%
    \centering
    \vspace{.5cm}
    \includegraphics[width=0.9\linewidth]{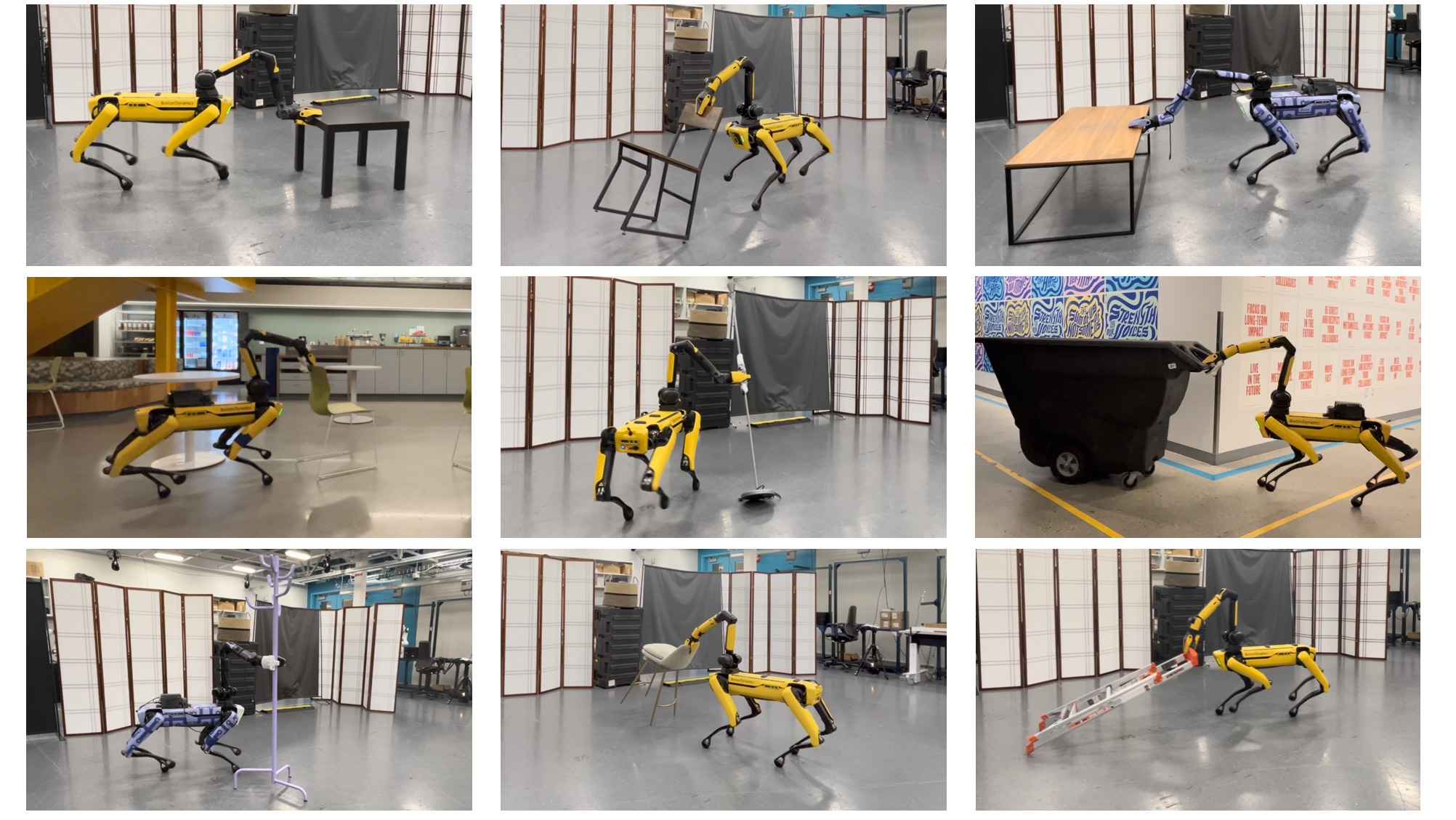}
    \captionof{figure}{\method \  enables robots to move a variety of large objects. }
    \label{fig:teaser}
    % \vspace{-.5cm}
    
}
\makeatother

\maketitle

\begin{abstract}
% Moving large objects, such as furniture, is a critical capability for robots operating in human environments. This task presents significant challenges due to two key factors: the need to synchronize whole-body movements to prevent collisions between the robot and the object, and the under-actuated dynamics arising from the substantial size and weight of the objects. These challenges also complicate performing these tasks via teleoperation. In this work, we introduce \method, a generalizable learning framework that leverages human-object interaction demonstrations to enable robots to perform large object manipulation tasks. Central to our approach is the Dynamic Chain, a novel representation that abstracts human-object interactions so that they can be retargeted to robotic morphologies. The Dynamic Chain is a spatial descriptor connecting the human and object root position via a chain of nodes, which encode the position and velocity of different interaction keypoints.We train policies in simulation using Dynamic-Chain-based imitation rewards and domain randomization, enabling zero-shot transfer to real-world settings without fine-tuning. Our approach outperforms both learning-based methods and teleoperation baselines across six evaluation metrics when tested on three distinct object types, both in simulation and on physical hardware. Furthermore, we successfully apply the learned policies to real-world tasks, such as moving a trash cart and rearranging chairs.

Moving large objects, such as furniture or appliances, is a critical capability for robots operating in human environments. This task presents unique challenges, including whole-body coordination to avoid collisions and managing the underactuated dynamics of bulky, heavy objects. In this work, we present RobotMover, a complete learning-based system for large object manipulation that leverages human-object interaction demonstrations to train robot control policies. RobotMover formulates the manipulation problem as imitation learning using a simplified spatial representation—referred to as the Interaction Chain—to capture essential human-object interaction dynamics in a morphology-agnostic way. We integrate this Interaction Chain into a reward structure and train policies in simulation using domain randomization to support zero-shot transfer to real-world hardware. The learned policies enable a Spot robot to manipulate various large objects—including chairs, tables, and standing lamp. Through extensive experiments across simulation and real-world platforms, we demonstrate that RobotMover achieves strong performance in terms of capability, robustness, and controllability, outperforming both learned and teleoperation baselines. Our system further supports practical applications by combining the learned policy with simple planning modules to accomplish long-horizon object transport and rearrangement tasks in the real world.
\end{abstract}


\section{Introduction}
\label{sec:introduction}

%%%%  Moving large object is inevitable, we can use robot to help.
\IEEEPARstart{M}{oving} large objects, such as furniture or appliances, is a common activity in human environments. Whether it involves rearranging a living room or organizing a workspace, these tasks often demand considerable physical effort. While people have traditionally handled such tasks manually, there is growing interest in developing autonomous robots to assist. Robotic systems capable of moving large objects could alleviate human physical strain and improve operational efficiency across a wide range of settings—from households to industrial logistics.

Despite significant progress in enabling robots to rearrange small items~\citep{SayCan, chi2023diffusion, black2024pi_0, team2024octo}, manipulating large objects presents a distinct set of challenges. These challenges stem from two key factors, as illustrated in Figure~\ref{fig:challenge}. First, large objects impose non-negligible spatial constraints: without sophisticated control policies, the object may collide with the robot or nearby structures, potentially causing damage. Second, such objects often exceed the robot’s lifting capacity, requiring the robot to manage object momentum and overcome ground friction during movement. These factors necessitate dynamic control policies that synchronize whole-body motion with object dynamics, while also addressing the under-actuated nature of the task.

%%%% Learning based method, teleoperation hard and time consuming. Data limit
\replaced{While existing methods have made tremendous progress in enabling robots to acquire various manipulation skills, they are not well suited to address the complexity of large object manipulation.}{Several approaches have been proposed to address these challenges.} Model-based approaches rely on accurately modeling the dynamics of the robot, the object, and their interactions. However, capturing the full range of object interactions and dynamics is difficult, which limits the generalizability of these methods across diverse scenarios. Reinforcement learning enables robots to learn through trial and error, but it requires carefully crafted reward functions that can generalize across different object types and configurations. Behavior Cloning through teleoperation has shown success in many applications~\citep{MobileAloha, lu2024mobile}, but it depends heavily on skilled operators and incurs high data collection costs—especially when dealing with large, heavy objects. These limitations highlight the need for more scalable and generalizable learning frameworks specifically tailored for large object manipulation.

%%%%  Introduce our method
In this work, we propose \method, a framework that enables robots to learn large object manipulation policies by imitating human-object interaction demonstrations. The framework leverages human and object movements as references to construct an imitation reward. Using this reward, we train a robot control policy in a simulation environment with domain randomization, enabling zero-shot transfer to real-world settings. \replaced{However, due to the  morphological differences between humans and robots, human demonstrations cannot be directly used as references for robot learning.}{A key aspect of this approach is to overcome the morphological differences between humans and robots, which makes it infeasible to directly copy the joint angles for imitation. } To address this challenge, we introduce the \replaced{Interaction Chain}{Dynamic Chain}, a low-dimensional, morphology-agnostic representation that captures the essence of agent-object interactions. The Interaction Chain is a Chain structure connecting keypoints from the agent’s root to the object’s root, with each keypoint serving as a node along the chain. The movement of the Interaction Chain captures how force is transmitted from the agent’s "core" to the object. \replaced{By using the Interaction Chain as the imitation reference, we eliminate the need for manually defining whole-body correspondences between the human and the robot, enabling more scalable and transferable policy learning.}{Compared to whole-body movements, the Dynamic Chain provides an abstract representation that can be easily retargeted to the robot's morphology. We use the Dynamic Chain as the imitation reference to train the robot's policy, enabling it to effectively manipulate large objects.}

%%%%  Experiment details
\replaced{We evaluate the proposed framework through a series of experiments conducted on the Spot robot platform in both simulation and real-world environments. Our results demonstrate that the learned policy can manipulate target objects while accurately tracking diverse desired velocities.}{To evaluate the effectiveness of our method, we conduct a series of experiments in both simulated and real-world environments.} In simulation, we compare our method against four learning-based baselines: reinforcement learning, end-effector tracking, \replaced{motion retargeting}{inverse kinematics methods}, \added{and Interaction Graph-based methods}. The results indicate that our approach consistently outperforms these baselines in tracking target object velocities. On hardware, we compare our policy against two learning-based baselines and two types of teleoperation methods. We evaluate performance across three key aspects—robustness, capability, and controllability—using six metrics in total. RobotMover outperforms all baselines across these metrics, highlighting its adaptability and effectiveness in handling diverse large objects. \replaced{Finally, we demonstrate the practical utility of our system through two real-world applications: trash cart transportation and chair rearrangement, achieved by combining the learned policy with different motion planners. }{Finally, by combining our learned policy with a motion planner, we demonstrate our framework in two real-world applications: trash cart transporation and chair rearrangement, highlighting the practical utility of our approach.}

%%%% Summarize contribution
In summary, this paper makes the following contributions:
\begin{itemize}
    \item \textbf{A new robot task:} We formalize \emph{large object manipulation} as an important and challenging robotic problem, characterized by the need to synchronize whole-body motion and develop dynamic control strategies to manage the shape, inertia, and frictional properties of large, heavy objects—challenges that are not typically encountered in small object.
    
    \item \textbf{A \deleted{learning} framework for large object manipulation:} We present a \deleted{learning} framework that enables robots to manipulate large objects by imitating the \emph{Interaction Chain}, a low-dimensional, morphology-agnostic representation of human-object interactions. \added{The method also provides practical guidelines for coordinating whole-body motion and handling complex object dynamics.}

    \item \textbf{Extensive validation:}  \replaced{We conductcomprehensive in both simulation and on physical hardware, demonstrating that our approach enables robust, generalizable large object manipulation across a variety of tasks and object types, and highlighting its applicability to real-world deployment.}{We conduct several experiments to evaluate our approach in simulation and hardware, demonstrating that our framework enables robots to reliably move diverse objects and can be used for real-world applications.}
\end{itemize}

\setcounter{figure}{1}
\begin{figure}[t]
\centering
\includegraphics[width=.995\linewidth]{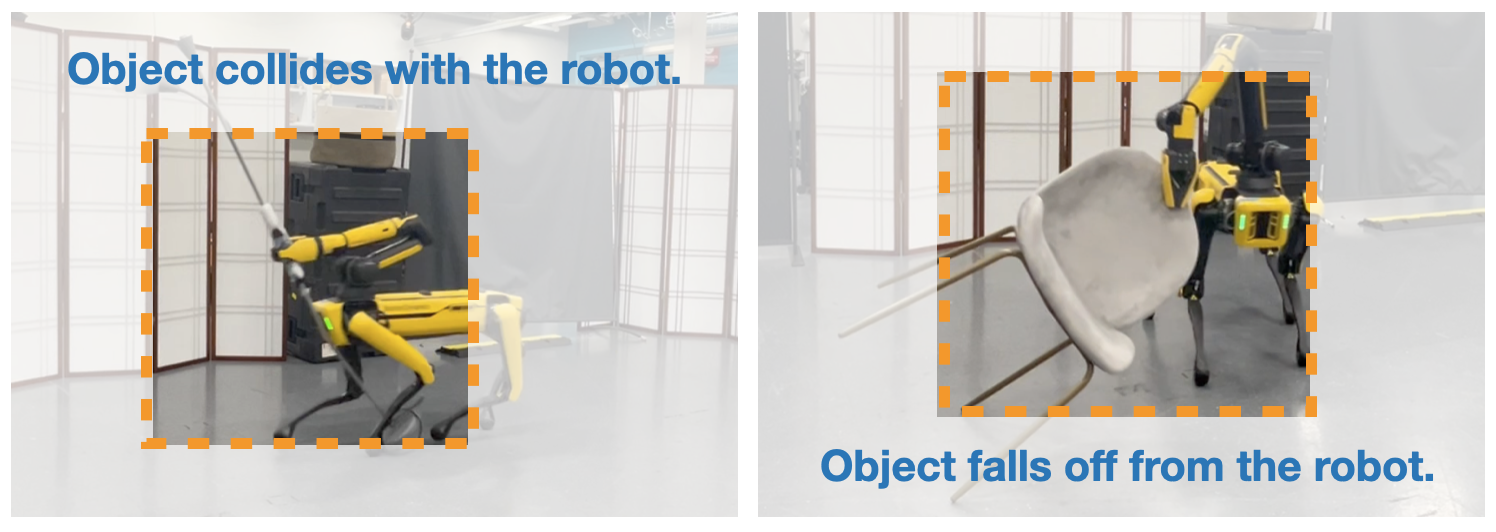}
\caption{Challenges of moving large objects include but not limit to object colliding with the robot and object falls off from the robot's gripper due to high momentum.}
\label{fig:challenge}
\vspace{-0.5cm}
\end{figure}

\added{The remainder of the paper is organized as follows. Section~\ref{sec:related_work} reviews related work in mobile manipulation, large object manipulation, and imitation learning. Section~\ref{sec:preliminary} introduces the preliminary background on imitation learning. Section~\ref{sec:method} describes the proposed RobotMover framework, including the Interaction Chain representation and the reward design. Section~\ref{sec:system} details the system implementation and policy architecture. Section~\ref{sec:experiments} presents experimental results in both simulation and hardware, followed by real-world applications. Finally, Section~\ref{sec:limitation} discusses the limitations of the current approach and directions for future work, and Section~\ref{sec:conclusion} concludes the paper.}

\section{Related Works}
\label{sec:related_work}

\subsection{Mobile Manipulation}

Mobile manipulation that involve robot locomotion and manipulation capabilities have been a prominent topic in robotics research~\citep{MobileAloha, TidyBot, Naoki_ASC, Behavior, Habitat, mittal2022articulated, DeepWholeBody, RT1, SayCan}. Some works tackles this problem using model-based control, which generally requires domain expertise to model the dynamic of the specific systems and tasks~\citep{krotkov2018darpa, wyrobek2008towards, khatib1996force, garrett2021integrated}.Recently, learning-based approaches have been applied to mobile manipulation, alleviating much of the heavy engineering effort. These methods have addressed a variety of challenging real-world tasks, such as mobile pick-and-place~\citep{Naoki_ASC, sun2022fully, Homerobot, ha2024umi}, manipulation of articulated objects~\citep{xiong2024adaptive, shafiullah2023bringing, yang2023moma}, and long-horizon cooking tasks~\citep{MobileAloha}. There has also been an interest in creating more interactive and user-friendly robotic systems by integrating primitive mobile manipulation skills with Large Language Models~\citep{TidyBot, RT1, SayCan}. However, most prior work has focused on robots interacting with objects that are small and light, where the dynamic effects of these objects are usually negligible for manipulation. In this work, we study robot learning for rearranging large and heavy objects, such as furniture or a trash truck, where it is necessary to account for these effects.

\subsection{Large Object Manipulation}

While robots capable of manipulating large objects are common in industrial applications, \replaced{they typically operate in structured environments and rely on large, powerful industrial arms}{they operate in controlled environments, where the objects are known and the dynamics more easily predictable}. Some efforts have been made to enable robots to move large objects in more unstructured settings, either by training policies through imitation of teleoperation demonstrations~\citep{MobileAloha} or via curiosity-driven exploration methods~\citep{mendonca2024continuously}. However, collecting teleoperation data is time-consuming and often tightly coupled to the specific robot platform, limiting generalization. Curiosity-driven methods, while flexible, face challenges in learning robust and reusable policies suitable for real-world deployment. Ravan et al.~\citep{ravan2024combining} \replaced{address the problem of manipulating a wheeled chair in complex environments; however, the relatively simple dynamics of a wheeled chair allow the robot to move the object without learning dynamic manipulation skills, and their work primarily focuses on high-level navigation planning rather than low-level manipulation control. In contrast, our objective is to train dynamic manipulation policies that enable robots to move large objects with complex, under-actuated dynamics in unstructured environments.}{addresses the problem of manipulating a wheeled chair without simulation. In this work, we introduce a generalizable learning framework that enables robots to manipulate large objects from human demonstrations.}

\subsection{Imitation Learning for Robotics}
%% rewrote the entire one
\added{Imitation learning is a framework that utilizes demonstration data to help robots acquire new skills. It has proven to be an effective approach for enabling robots to learn a wide range of capabilities, from sophisticated manipulation to agile locomotion~\cite{MobileAloha, chi2023diffusion, jang2022bc, li2023crossloco,  shafiullah2022behavior, li2023ace, smith2023learning}. Imitation learning methods are typically categorized into two main paradigms: behavioral cloning and reinforcement learning .}

\added{In the behavioral cloning setting, it is assumed that the robot and the demonstrator share the same action space. Under this assumption, expert demonstrations can be directly used as supervised training data, allowing a robot policy to be learned via standard supervised learning. Demonstration data can be sourced from fixed datasets~\cite{vuong2023open, MobileAloha, chi2024universal} or actively collected through interactive methods such as Dataset Aggregation (DAgger)~\cite{li2019using}.}

\added{In the reinforcement learning setting, demonstration data do not provide direct action supervision. Instead, demonstrations are used to define a reward function that encourages the robot to imitate the expert’s behavior. For example, Peng et al.~\cite{RoboImitationPeng20, peng2018deepmimic} leverage motion capture data to construct imitation-based rewards, and the robot policy is trained to maximize these rewards. Other works~\cite{peng2021amp, peng2022ase} employ adversarial imitation learning, where a discriminator is trained to distinguish between expert and robot behaviors, and the robot learns to generate motions that are indistinguishable from expert demonstrations. To learn multi-agent interaction skills, Zhang et al.~\citep{InteractionGraph} introduce Interaction Graphs as a representation for modeling interaction and using them to learn humanoid robot interaction behaviors from human demonstrations. In this work, we address the problem of robot learning for large object manipulation by imitating human-object interactions. In our setting, it is critical to consider both the coordinated movements of the human and the object, as well as the significant morphological differences between humans and robots.}

\section{Preliminary}
\label{sec:preliminary}

\subsection{Interaction Graph}
\added{Interaction Graphs~\citep{InteractionGraph} are a representation designed to capture the spatial and dynamic relationships between multiple interacting entities, such as human-to-human or human-to-object interactions. In Interaction Graphs, nodes typically correspond to keypoints or joints of the interacting bodies, while edges encode pairwise spatial relationships or contact information. This graph structure provides a compact yet expressive summary of the interaction dynamics, which can be used as a reference for imitation learning.}

\added{Interaction Graphs have proven effective for tasks involving humanoid robots or agents with similar morphologies to humans, where rich body-object contact patterns need to be preserved. However, the high dimensionality and dense connectivity of the graph can pose challenges when transferring interaction knowledge to robots with significantly different morphologies. In particular, adapting the graph structure to different embodiment types can require careful redesign or simplification, limiting its scalability.}

\added{In this work, we draw inspiration from Interaction Graphs but propose a simpler and more generalizable representation—the Interaction Chain—that better accommodates embodiment differences while still capturing the essential robot-object interaction dynamics.}

\subsection{Imitation Learning}
Our robot learning framework is built upon the imitation learning paradigm, which enables robots to acquire skills by learning from demonstration data. The objective is to teach the robot how to move large objects by replicating human-object interaction behaviors. We begin with a demonstration trajectory $\xi = {p^{ho}_1, p^{ho}_2, \dots, p^{ho}_T }$, where each aggregated pose $p^{ho}_t = (p^h_t, p^o_t)$ represents the human pose $p^h_t$ and the object pose $p^o_t$ at time step $t$. These trajectories, typically collected via motion capture systems or online datasets, serve as the basis for defining a reward signal that evaluates how well the robot replicates the demonstrated behavior.

The imitation learning process is formulated as a reinforcement learning (RL) problem, where the robot learns a policy $\pi(a_t | s_t)$ to maximize the expected cumulative reward: \begin{align} J(\pi) = \mathbb{E}\left[\sum_{t=1}^T r(s_t, a_t)\right], \end{align} where $r(s_t, a_t) = r^{imi}_t + r^{reg}_t$ consists of an imitation reward $r^{imi}_t$ and a regularization reward $r^{reg}_t$. The imitation reward encourages the robot to produce behaviors that closely align with the demonstrations, while the regularization reward promotes smooth and efficient motions.

In many standard imitation learning settings, the demonstrator and the robot share similar morphologies, which enables straightforward reward designs, such as direct pose or velocity matching. However, in our case, the human demonstrator and the robot possess significantly different morphologies, making it non-trivial to apply simple imitation reward schemes. This necessitates the design of morphology-agnostic representations and reward functions that can bridge the embodiment gap.

\section{RobotMover}
\label{sec:method}

\begin{figure}[t]
\centering
\includegraphics[width=.995\linewidth]{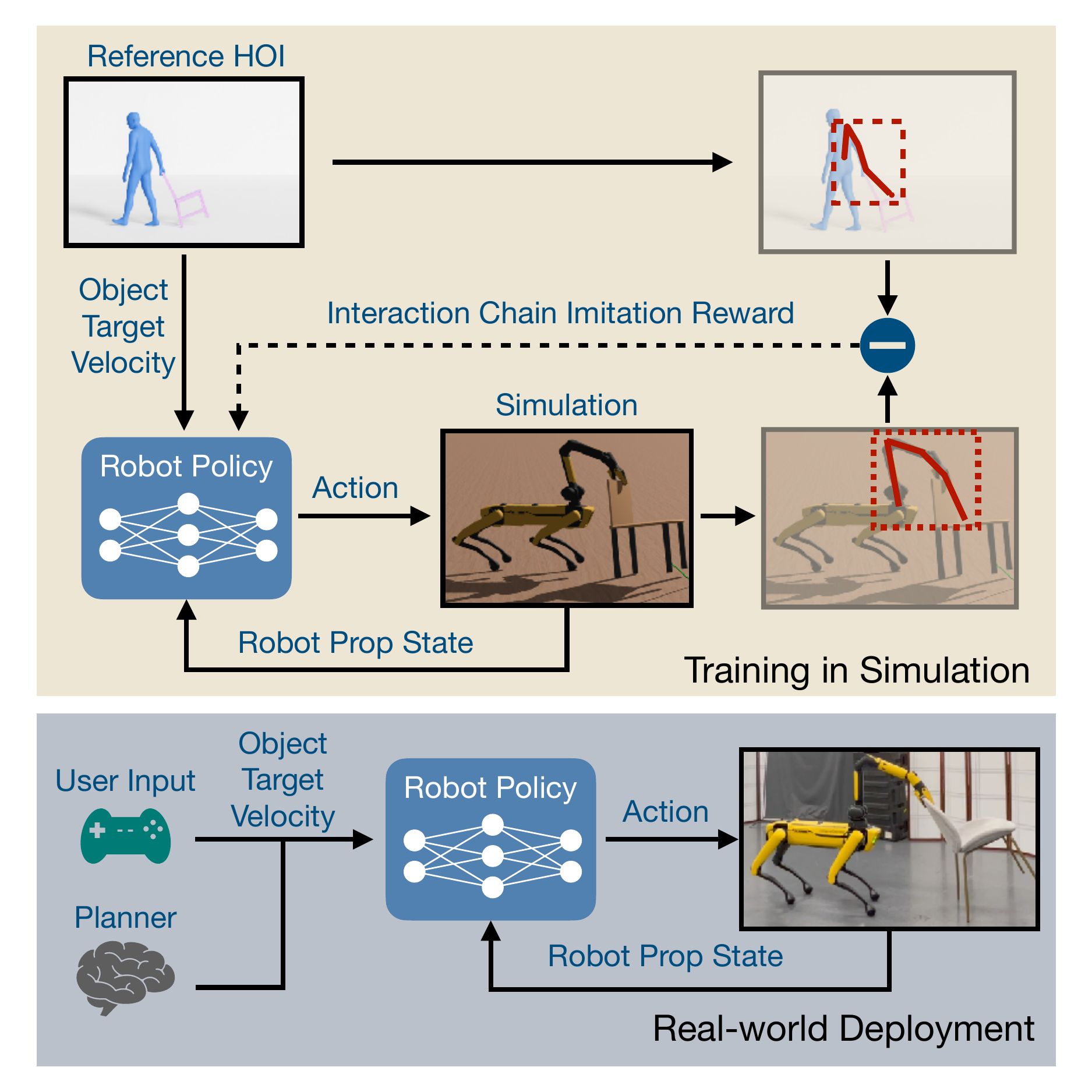}
\caption{Method Overview. RobotMover enables robots to learn to move large objects by imitating human-object interaction demonstrations. The framework leverages a novel representation, the Dynamic Chain, which captures the interaction dynamics between the agent and object while remaining agnostic to the agent’s embodiment. This representation is used to design an imitation reward that guides the robot’s imitation learning process.}
\label{fig:overview}
\vspace{-0.5cm}
\end{figure}

\added{We propose RobotMover, a learning framework that enables robots to learn large object manipulation by imitating human-object interaction demonstrations. An overview of the framework is shown in Fig.~\ref{fig:overview}. RobotMover formulates the problem as learning a robot control policy that moves the object to track target object velocities through imitation. To bridge the embodiment gap between humans and robots, we introduce a morphology-agnostic representation, the Interaction Chain, which captures the essential agent-object interaction dynamics and supports the design of an imitation reward. This representation allows the robot to learn from human demonstrations without requiring direct body-to-body correspondence. In the following subsections, we first elaborate the problem formulation, then introduce the Interaction Chain representation, describe how to extract the Interaction Chain from both human demonstrations and robot interactions, and finally present the reward design based on the Interaction Chain.}

\subsection{\replaced{Problem Formulation}{Cross-embodiment Imitation Learning}}

\deleted{While imitation learning is effective for transferring skills, the dynamic and morphological differences between humans and robots pose significant challenges. These differences, such as variations in degrees of freedom and body structure, prevent direct replication of human movements. To address this, we utilize \textit{cross-embodiment imitation learning}, which leverages a shared representation to align the interaction dynamics between human and robot.}

\deleted{The imitation reward function is designed to measure the alignment between the human and robot interaction dynamics:}

\deleted{A critical component of cross-embodiment imitation learning is the design of appropriate shared representations and projection functions. The representation must be concise enough to be reconstructed from both robot and human poses, while also being expressive enough to capture the gist of the behavior.}

Our objective is to learn a robot control policy that enables a robot to manipulate large objects by controlling both the robot’s motion and the resulting object movement. Specifically, the policy takes as input the target velocity of the object and outputs a control signal that drives the robot’s behavior to achieve the desired object motion. By adjusting the target velocity, a user or a high-level planner can command different manipulation goals, such as moving an object to a specific location or reorienting it to a desired heading.

To train such a policy, we leverage human-object interaction motion capture demonstrations of object manipulation. These demonstrations provide examples of coordinated human-object movements that can serve as references for the robot to imitate. However, a key challenge arises: due to the dynamic and morphological differences between humans and robots, direct replication of human movements is infeasible. Humans and robots often differ in their degrees of freedom, body structure, and actuation capabilities, complicating the transfer of demonstrated behaviors.

To address this issue, we adopt a cross-embodiment imitation learning approach. The core idea is to map both human and robot poses into a shared representation space that captures the essential interaction dynamics while abstracting away embodiment-specific details. The imitation reward function is then defined based on the similarity between the human and robot representations: \begin{align} \label{eqn:imit_reward} r^{imi}_t \propto -| \Psi^{ho}(p^{ho}_t) - \Psi^{ro}(p^{ro}_t) |, \end{align} where $\Psi^{ho}$ and $\Psi^{ro}$ are projection functions that embed the human aggregated pose $p^{ho}_t$ and the robot aggregated pose $p^{ro}_t = (p^r_t, p^o_t)$ into the shared space. This formulation enables the robot to evaluate and imitate human-object interactions without requiring direct body-to-body correspondence.

A critical component of this framework is the design of appropriate shared representations and projection functions. The representation must be concise enough to be reconstructed from both human and robot poses, while expressive enough to capture the essential behaviors necessary for successful large object manipulation.

\subsection{Interaction Chain}
\label{Method:Interaction_Chain}

\begin{figure}[t]
 \centering
\includegraphics[width=.995\linewidth]{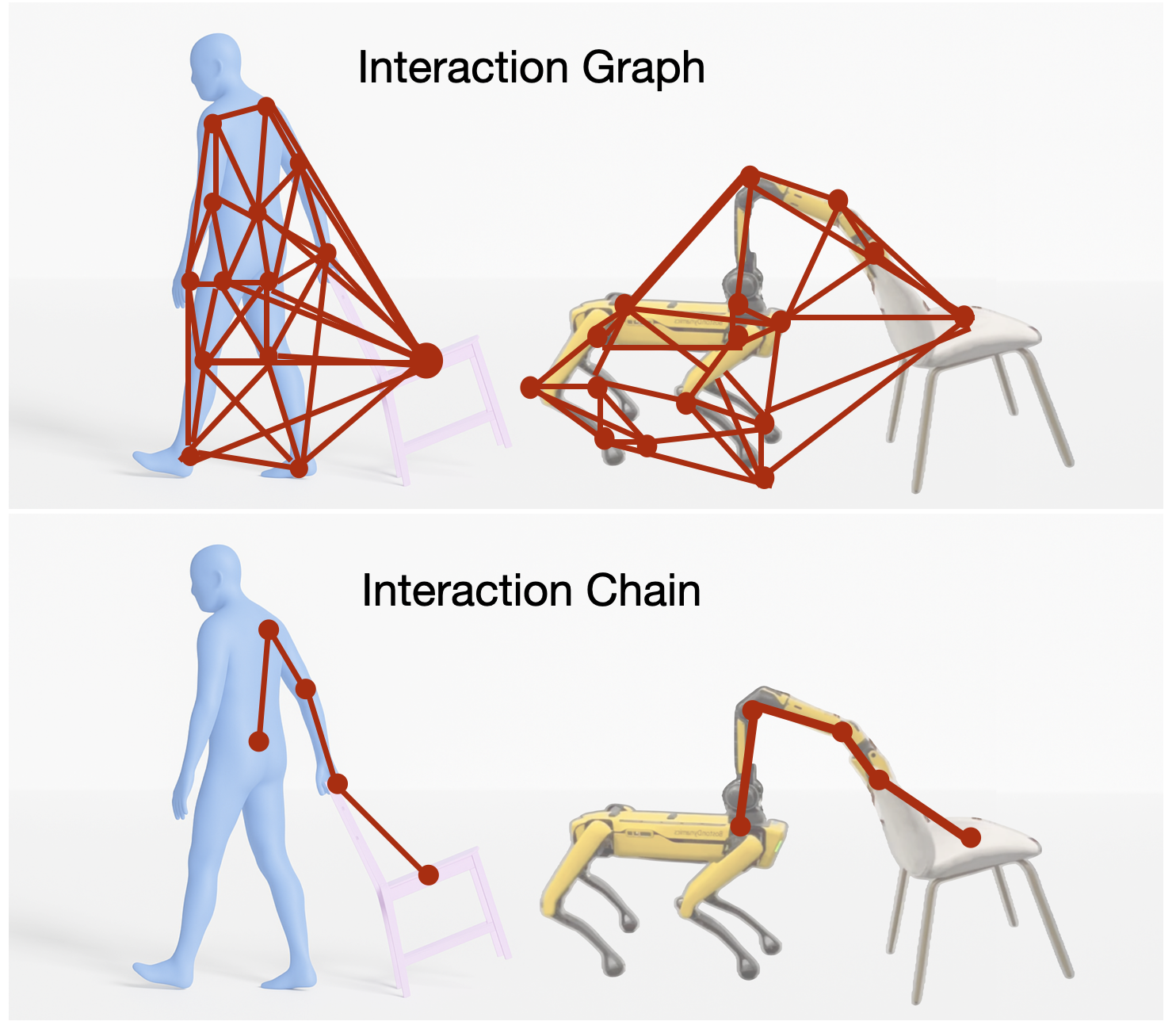}
 \vspace{-.5em}
\caption{Interaction Graph (top) vs. Interaction Chain (bottom). The Interaction Chain offers a simpler representation to describe agent-object interactions, which can better transfer to other morphologies. This allows us to use a human demonstration to guide the reward of a robot policy.  
% This allows it to more easily transfer to new domains, such as robot-object interaction, for reward design.
}
\label{fig:DC_vs_IG}
 \vspace{-0.2cm}
\end{figure}

To effectively capture the semantics of agent-object interactions while providing a meaningful metric to guide the robot’s learning process, we propose the Interaction Chain—a chain-based spatial descriptor inspired by prior works such as Interaction Graphs~\citep{InteractionGraph} and spatial relationship descriptors~\citep{ho2010spatial}. \added{The Interaction Chain encodes interaction information as a sequence of nodes connected by edges, offering a compact and interpretable representation of the dynamics involved.}

To construct the Interaction Chain, we place a set of markers on salient locations of both the agent and the object. One end of the chain is anchored at the object’s root, while the other end is anchored at the agent’s root. Intermediate nodes are placed at keypoints such as the shoulder, elbow, and the contact point between the agent and the object. The state of the agent-object Interaction Chain at time $t$ is defined as $c^{ao}_t = (x^{ao}_t, q^{ao,i}_t)$, where $x^{ao}_t$ represents the object’s position, orientation, and velocity in the global coordinate frame, and $q^{ao,i}_t$ denotes the orientation of the $i$-th chain segment in the object-plane coordinate. This arrangement captures how forces are transmitted from the agent’s core through the chain to move the object.

Figure~\ref{fig:DC_vs_IG} illustrates an example of the Interaction Chain, where edges and nodes connect the agent’s root to the object’s root via the agent’s body and contact point. The internal edges encode the agent’s body and arm configurations, while the final edge models the spatial relationship between the agent and the object. Thus, the Interaction Chain captures both absolute poses and the dynamic coordination between the agent and the object.

Compared to prior representations such as Interaction Graphs~\citep{InteractionGraph}, the Interaction Chain offers a lower-dimensional and simpler abstraction. Whereas Interaction Graphs often include exhaustive connections—such as linking the agent’s feet to the object—the Interaction Chain selectively captures only the most relevant interaction dynamics. This compactness improves transferability across different morphologies and facilitates learning more generalizable manipulation skills.

\subsection{Building Interaction Chain from Human Demonstrations}

\begin{figure}[t]
   \centering
  \includegraphics[width=.95\linewidth]{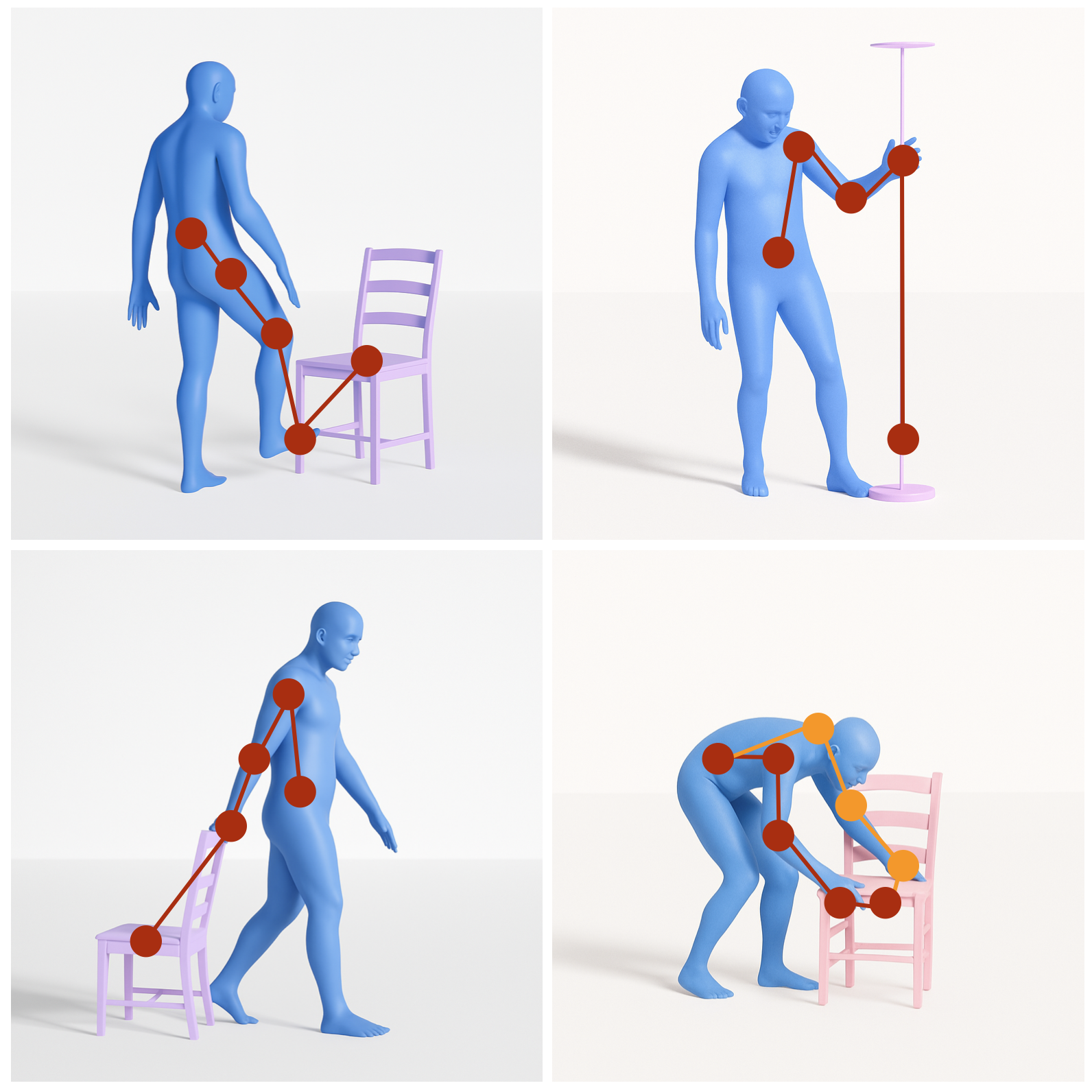}
   \vspace{-.5em}
  \caption{ \added{Different human-object interaction strategies lead to different Interaction Chains.} }
  \label{fig:ICs}
   \vspace{-0.2cm}
  \end{figure}

\added{Human demonstrations of large object manipulation can vary depending on which body parts primarily engage with the object. To accurately capture the key interaction dynamics, we construct the Interaction Chain based on the dominant agent-object contact observed in each demonstration. Fig.~\ref{fig:ICs} presents different examples of Interaction Chains.}

\added{For demonstrations where a single arm dominates the interaction, the Interaction Chain is constructed by connecting the object’s root to the agent’s root through keypoints along the active arm. Markers are placed along salient joints such as the wrist, elbow, and shoulder, forming a sequential chain that reflects how forces are transmitted from the agent to the object. If the left arm is primarily responsible for the manipulation, the chain is constructed through the left-side joints; if the right arm is dominant, the chain follows the right-side joints. In demonstrations involving substantial use of both arms, we extract chains along each arm separately. In this work, we primarily focus on single-arm manipulation scenarios, where the Interaction Chain provides a concise and consistent representation of the demonstrated behavior.}

\subsection{Building Robot-Object Interaction Chain}
\added{Similar to human demonstrations, the robot’s Interaction Chain is dynamically constructed based on the physical contact between the robot and the object. During training and deployment, we do not predefine which end-effector or body part the robot should use to manipulate the object. Instead, the Interaction Chain is formed automatically once contact is established. Specifically, when the robot makes contact with the object, we construct a chain that connects the object’s root to the robot’s root through the contacting body parts. This dynamic formation allows the Interaction Chain to flexibly capture a variety of contact strategies without requiring manual specification of the manipulation limb.}

\subsection{Reward Design}
The core of our reward design is based on the similarity between the Interaction Chain \added{extracted from the human demonstration and the Interaction Chain generated by the robot during manipulation}. Specifically, let $c^{ho}_t = \Psi^{ho}{dc}(p^{ho}_t)$ denote the human demonstration’s Interaction Chain and $c^{ro}_t = \Psi^{ro}{dc}(p^{ro}_t)$ denote the robot’s Interaction Chain at time $t$. The imitation reward is derived by measuring the distance between these two chains, computed as the sum of differences between the global positions of each corresponding node on the normalized chains:
\begin{align} 
   \label{eqn:norm_error} 
   err^c_t = ||x^{ho}_t - x^{ro}_t|| + \sum_{i=0}^{N-1} \alpha^i \| q^{ho,i}_t - q^{ro,i}_t \|, 
\end{align} 
where $|x^{ho}_t - x^{ro}_t|$ measures the difference between the object movements induced by the human and the robot, and $\alpha^i$ represents a weighting factor applied to each chain segment with i=0 indicates object root. We assign higher weights to segments closer to the object to emphasize precise tracking of critical interaction points. \added{Here, $N$ denotes the number of nodes between the robot's and the human's Interaction Chains. In our experiments, the robot and human have the same node number. In the scenarios when robots and demonstrator have different node number, we can use the smaller number of the two chains as $N$ or use dynamic wrapper to measure the chain difference. }

However, if no contact occurs, no Interaction Chain can be formulated on the robot side, and thus no imitation reward is computed. Formally, the imitation reward is defined as: 
\begin{align} 
   \label{eqn:imi_2} 
   r^{imi}_t = e^{-err^c_t} \times \mathbf{1}(f^{ee} > 0), 
\end{align} 
where $f^{ee}$ indicates the contact force magnitude at the robot’s end-effector, and $\mathbf{1}(\cdot)$ is the indicator function. This design ensures that the robot is only rewarded for coordinated manipulation behaviors that maintain physical interaction with the object.

In addition to the imitation reward, we incorporate auxiliary regularization terms to improve motion quality, including height and torque penalties, body rotation penalties, and action rate penalties following~\citep{rudin2022learning}. These terms encourage smoother and more efficient robot movements, mitigate abrupt or unstable actions, and ultimately improve the robustness and naturalness of the learned manipulation behaviors.

\section{\replaced{System Overview}{Model Representation}}
\label{sec:system}

\added{We use the Boston Dynamics Spot robot, a quadrupedal robot equipped with a robotic arm mounted on its body as our platform. It features three actuators per leg and a seven-degree-of-freedom robotic arm, including a gripper motor, resulting in a total of 19 motors. The robot's default height is 610~mm and the robot arm's extension up to 984~mm, allowing it to manipulate large objects. The robot control policy operates at 20 Hz and is trained in simulation using Proximal Policy Optimization (PPO)~\citep{PPO}. During training, 4096 environments are simulated in parallel on a single NVIDIA GeForce RTX 3080 Ti GPU for a period of about 4 hours.  For hardware experiments, we deploy the simulation-trained policy directly without additional fine-tuning. In the following subsections, we first describe the human-object interaction demonstration dataset used for training, then present the simulation environment setup, and finally elaborate on the model representation adopted in our framework.}

\subsection{Demonstration Dataset}
For demonstration data, we utilize the OMOMO dataset~\citep{OMOMO}, a high-quality motion capture dataset of human-object interactions that provides precise motion trajectories for both humans and objects. The dataset includes all object types involved in our experiments. To train a policy for a specific object type, we aggregate all relevant demonstrations. For example, when training a policy for moving chairs, we include demonstrations of humans manipulating both wooden chairs and white chairs from the OMOMO dataset. The resulting demonstration collections contain diverse human motion trajectories for each object category. In our experiments, most demonstrations are human moving object using single arm and we include a brief analysis of policy training using two-arm demonstrations is provided in Sec.~\ref{sec:2-arms}. Additional details about the dataset composition are available in the Appendix.

\subsection{Simulation Setup}
\label{sec:simulation_setup}
Our simulation environment is built on the Genesis simulator~\citep{Genesis}. We construct two types of simulation environments: a simplified dynamics environment and a full dynamics environment.

The simplified dynamics environment is designed to align with the hardware interface, where we do not have direct control over Spot’s leg motors. In this setup, the robot policy outputs the robot’s root linear and angular velocities in the robot’s local coordinate frame, along with the desired arm poses. This results in a 13-dimensional action space (6 dimensions for root velocities and 7 for arm poses). At each environment step, the robot’s root velocity is set according to the policy output, and the physics simulation proceeds with the desired arm pose as a target.

The full dynamics environment, in contrast, is used to validate our method without the limitations imposed by hardware constraints. In this setup, the policy controls all motors directly, leading to a 19-dimensional action space, with each dimension corresponding to the desired pose of an individual joint motor.

For the objects in the simulation, we construct custom object models using primitive geometries to approximate those provided in the OMOMO dataset. This approach ensures that the final object shapes remain similar to the dataset objects while significantly improving computational efficiency. Using the original high-fidelity mesh files from OMOMO, which contain complex triangular meshes, would increase computational overhead during object-robot interactions, particularly under contact dynamics.

When resetting the simulation environment, the robot’s xy-position is set to the origin, and the object is placed at a position and orientation aligned with the reference demonstrations. To introduce variability, we randomize the robot’s initial height, as well as the object’s initial position, orientation, mass, and friction coefficient. We use a single geometric model per object type, meaning that the object's shape is not randomized. This decision is motivated by two key factors. First, in our simulation setup, all objects must be loaded at the start of the simulation, and during each simulation step, the contact state of every object is computed—even if only one object is manipulated in a given episode—making geometric randomization computationally expensive. Second, we argue that geometric differences are ultimately reflected in dynamic variations once the object is grasped, and these dynamics can be effectively randomized through variations in friction, mass, and initial poses, without requiring explicit shape randomization.

\subsection{Model Representation}
The robot policy’s observation space consists of two main components: the object’s target velocity and the robot’s proprioceptive data. The object’s target velocity is derived from human demonstrations during training and can be provided either by a human operator or generated by a high-level planner during test time.

The proprioceptive data are acquired from the robot’s onboard sensing system. Notably, the observation space excludes direct object state information—such as the object’s orientation, velocity, or visual input—thereby facilitating sim-to-real transfer, improving generalization across object variations, and enabling deployment without requiring external object sensing.

Separate policies are trained for different categories of objects to accommodate their unique dynamic properties. The policy observation space includes: 
\begin{itemize} 
    \item The object's target velocity (3 dims) 
    \item The robot's root linear and angular velocities (6 dims) 
    \item The robot's joint angles (18 dims) 
    \item The robot's joint velocities (18 dims) 
    \item The robot's gripper open angle (1 dim) 
    \item The robot's gripper contact with the object (1 dim) 
\end{itemize} 
Here, the object's target velocity is expressed in the current object-plane frame, while the robot’s root velocities are measured in the robot-plane coordinate frame.

We built two types of simulated environments: a simplified dynamics (SD) environment and a full dynamics (FD) environment, as described in Sec.~\ref{sec:simulation_setup}. These settings correspond to different policy action spaces: \begin{itemize} 
    \item SD Env: the robot's root velocities and arm joint angles (13 dims) 
    \item FD Env: the robot's desired joint angles (19 dims) 
\end{itemize}

To train the critic network, we augment the policy observation with additional object-related information: 
\begin{itemize} 
    \item The object's target orientation-global frame (4 dims) 
    \item The object's current orientation-global frame (4 dims) 
    \item The object's root linear and angular velocities (6 dims) 
\end{itemize}

Both the policy and the critic are implemented as fully connected neural networks with three hidden layers, consisting of [512, 256, 128] units, respectively, using ELU activation functions~\citep{ELUs}.

\begin{figure}[t]
\label{fig:simulation_setting}
\centering
\includegraphics[width=.995\linewidth]{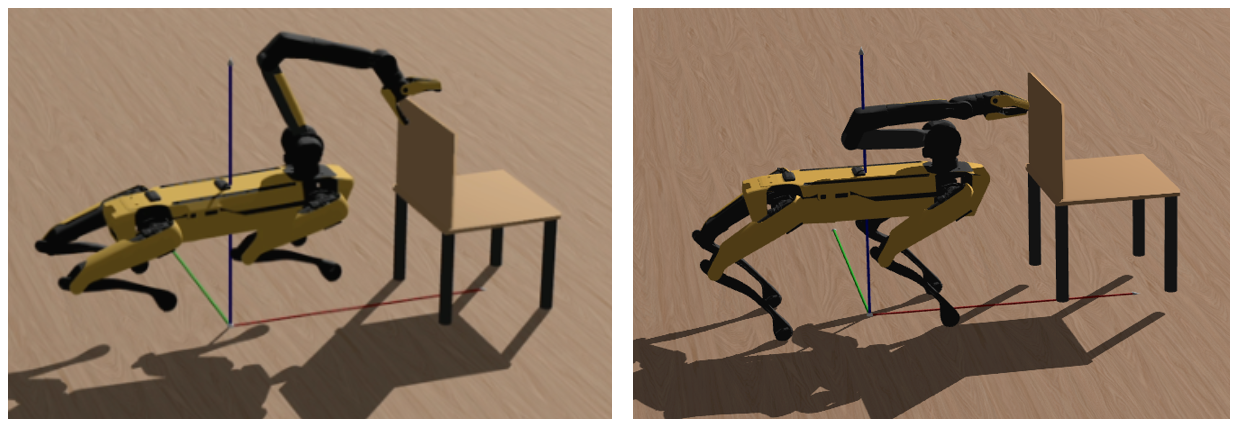}
\vspace{-.5em}
\caption{Two simulation settings. Left: simple dynamic environment, without legged control. Right: fully dynamic environment, controlling all the robot's motors.}
 \vspace{-0.2cm}
\end{figure}

\section{Experiments}
\label{sec:experiments}
Our experiments aim to answer the following questions:
(1) Can robots learn to manipulate large objects using \method?
(2) How does \method\ compare to existing learning-based and teleoperation methods?
(3) Can the learned policies be effectively applied to real-world tasks?

To address these questions, we conduct a series of experiments in both simulation and real-world settings. The experiments involve three categories of objects—chairs, tables, and standing racks—with variations in weight and surface texture within each category. We compare \method\ against a range of baseline methods, including learning-based and teleoperation approaches, using multiple evaluation metrics. We also assess the real-world applicability of the learned policies by deploying them in practical manipulation tasks. In addition, we investigate the potential of leveraging two-arm demonstrations to support robot learning of large object manipulation.

This section is organized as follows. We first present the simulation setup and results, followed by the hardware setup and experimental findings. We then demonstrate the deployment of the learned object manipulation policies in real-world scenarios. Finally, we explore two-arm demonstrations as a source of training data.Our results can be best watched in the supplementary video.

\subsection{Simulation Experiments Setup}
\label{sec:exp_sim_setup}

\begin{figure*}[t]
\centering
\includegraphics[width=.99\linewidth]{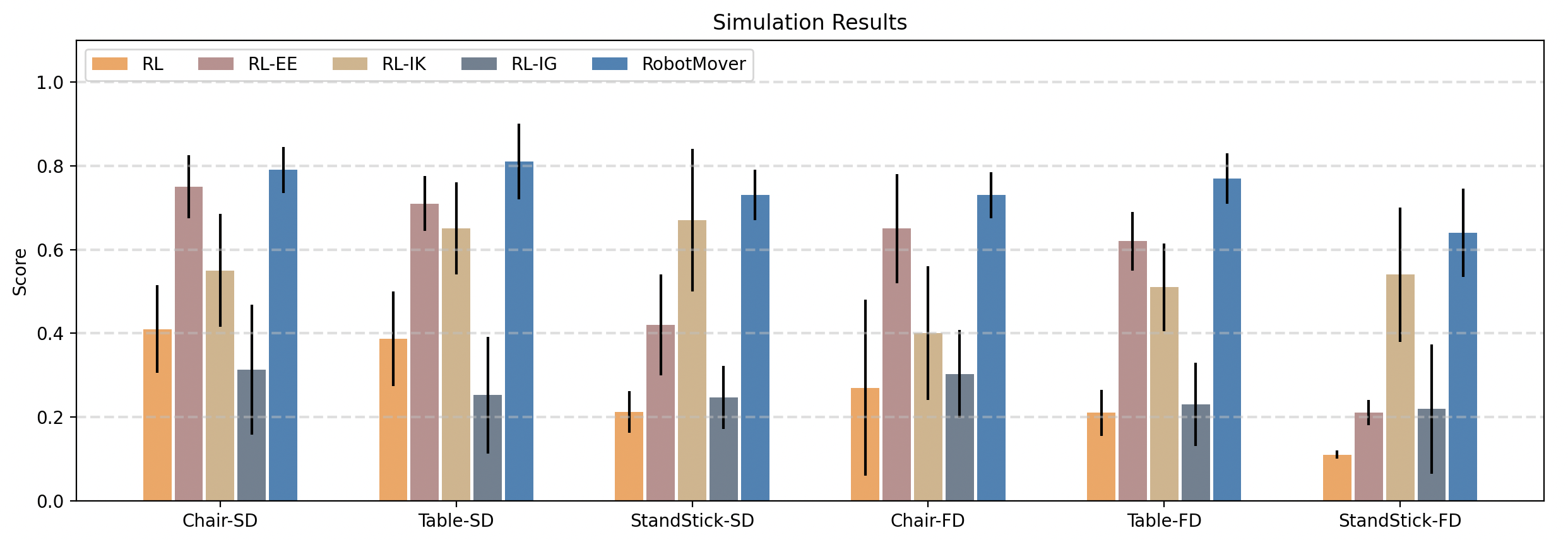}
\caption{Quantitative results of simulation experiments. Here, `SD' represents simple-dynamics environment while `FD' indicates full-dynamics environment. Our result indicates \textbf{\method} outperform baseline methods in almost every settings.}
\vspace{-0.5cm}
\label{fig:simulation_barplot}
\end{figure*}

Five methods are involved in the simulation experiment, with four baselines methods and our proposed \method:

\begin{itemize} 
\item \textbf{RL}: A baseline that uses reinforcement learning to train a robot control policy for manipulating the target object. The reward function is defined based on the difference between the object’s root trajectory and the reference trajectory.

\item \textbf{RL-EE}: This baseline incorporates the demonstrator's end-effector position in the global frame alongside the object’s root trajectory as imitation goals. Compared to RL, RL-EE uses the end-effector position as a heuristic to guide exploration and aid policy convergence.

\item \textbf{RL-IK}: Similar to RL-EE, this baseline incorporates the demonstrator’s end-effector position, but measured in the robot's local frame. This provides localized guidance during policy training by referencing both the object’s root trajectory and the end-effector’s position relative to the robot.

\item \textbf{RL-IG}: \added{This baseline utilizes the Interaction Graph (IG) representation from demonstrations for robot policy learning. To use the IG as a reference, we design mapping and normalization functions between the 253-edge human-object interaction graph and the 171-edge robot-object interaction graph. For practical implementation in simulation, we simplify this by selecting and aligning 15 key edges from the full interaction graph to serve as guidance for policy learning.}

\item \textbf{\method(ours)}: \method\ leverages both the object’s movement trajectory and the demonstrator’s body movement, captured through the Interaction Chain, to guide policy learning. This provides rich interaction information, enabling the robot to acquire robust object manipulation skills.

\end{itemize}

These methods are evaluated based on their ability to track reference object trajectories. The tracking score $v^{track}$ is defined as: 
\begin{align} 
    \label{eqn:norm_error_metric} 
    v^{track} = \frac{1}{T}\sum_{t=0}^{T} e^{-|x^{ho,xy}_t - x^{ro,xy}_t|}, 
\end{align} 
where $x^{ho,xy}_t$ and $x^{ro,xy}_t$ denote the 2D xy-positions and headings of the reference and experimental objects at time $t$, and $T$ represents the total number of time steps in an episode. The closer the object's trajectory is to the reference, the higher the tracking score. We evaluate each method using 5 random seeds, with performance measured over 10 randomly sampled reference trajectories per seed.

\subsection{Simulation Comparison Results}
Figure~\ref{fig:simulation_barplot} presents a quantitative evaluation of the simulation results. The results demonstrate that the proposed \method\ outperforms all baselines in both the simplified and full dynamics environments.The \textbf{RL} baseline exhibits the lowest average score with high variance. This can be attributed to the lack of heuristic guidance, causing the RL method to struggle in finding effective solutions and often leading to convergence at suboptimal local minima. The \textbf{RL-EE} baseline improves upon RL by providing the global position of the end-effector, corresponding to the robot-object contact point. In some chair and table experiments, \textbf{RL-EE} achieves results comparable to \textbf{\method}. However, since it lacks information about the robot’s full arm pose, the resulting policies often produce unnatural arm configurations that are difficult to transfer to hardware. We further discuss this issue in the hardware results section.The \textbf{RL-IK} baseline incorporates arm pose information by referencing the end-effector’s position in the robot’s local frame. While this provides localized guidance, it performs poorly in certain scenarios, such as chair manipulation, where \textbf{\method} achieves a score of $0.79 \pm 0.04$ compared to $0.55 \pm 0.06$ for \textbf{RL-IK}. This performance gap arises because the local-frame arm poses do not naturally translate into effective global motions, particularly due to morphological differences between the human demonstrator and the robot. As a result, the learned policy struggles to generalize to diverse target object velocities. \added{When comparing \textbf{RL-IG} to \method, we observe that \method\ achieves consistently higher performance across all settings. We attribute this to the Interaction Chain’s ability to capture key aspects of the interaction—object motion, contact points, and arm configuration—without relying on exhaustive graph matching. In contrast, the high-dimensional structure of the Interaction Graph can unnecessarily constrain the robot’s motion, particularly for body parts less critical to the manipulation task, such as leg poses or root orientation.}

\begin{figure*}[t]
    \centering
    \includegraphics[width=.99\linewidth]{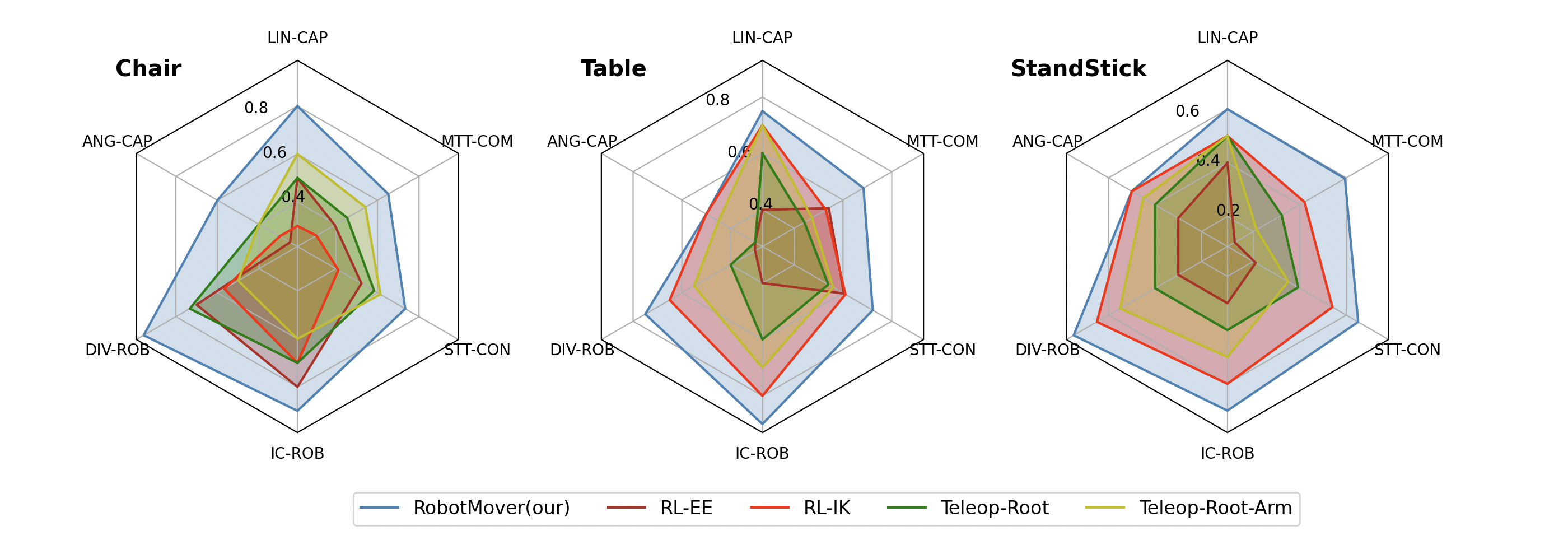}
    \vspace{-.5em}
    \caption{Summary of Hardware Quantitative Results. We compare our proposed \textbf{\method} with two learning-based baselines, \textbf{RL-EE} and \textbf{RL-IK}, as well as two teleoperation methods, \textbf{Teleop-Root} and \textbf{Teleop-Root-Arm}, across six evaluation metrics. The results show that \textbf{\method} outperforms all baseline methods in every evaluation.}
    \label{fig:hardware_results}
    \vspace{-0.5cm}
    \end{figure*}

\subsection{Hardware Experiments Setup} \label{sec:hardware_setup}

We conduct hardware experiments to evaluate the performance of our method in the real world. Experiments are performed using the Spot robot from Boston Dynamics. We deploy the policy trained in the simple dynamic simulation to the real-world robot without fine-tuning. To provide a comprehensive evaluation of our proposed method, we compare \method~against four baselines, including two learning-based methods and two teleoperation-based methods. These evaluations are conducted on three types of objects: chairs, tables, and racks, focusing on three aspects of the controller: capability, robustness, and controllability. In our hardware evaluation, unlike simulation where the robot is initialized at its default pose, we first teleoperate the robot to have a stable grasp with the target object and evaluate our method from that point, focusing on the performance of the `moving' part.

The two learning-based baselines are \textbf{RL-EE}, \textbf{RL-IK} introduced in section:~\ref{sec:exp_sim_setup}. The two teleoperation baselines are:

\begin{itemize}
    \item \textbf{Teleop-Root}: In this baseline, an operator teleoperates the robot by directly controlling its root linear and angular velocities, while the arm's movements rely on the robot's passive compliance for object interaction.
    \item \textbf{Teleop-Root-Arm}: Extending Teleop-Root, this baseline sets the robot arm's desired position to the initial contact pose, providing a more direct arm control during manipulation.
\end{itemize}

The evaluation focuses on three critical properties of the controller, each capturing a unique dimension of performance. \textbf{Capability (CAP)} measures the robot's ability to achieve maximum velocity while maintaining stability. \textbf{Robustness (ROB)} assesses the controller's performance across varying conditions and object dynamics. \textbf{Controllability (CON)} evaluates the precision and accuracy of trajectory tracking during task execution.

We define two metrics for each of the properties in a total of six metrics:

\begin{itemize}
    \item \textbf{Max Linear Velocity (LIN-CAP)}: Measures the maximum absolute linear velocity at which the robot can operate stably without the object falling off the gripper or colliding with the robot. Stability is defined as maintaining control for 8 seconds. Each method is evaluated over 5 runs, with success requiring at least 4 successful trials.We measured the maximum linear velocity, starting at 0.5 m/s and adjusting by $\pm0.05$ m/s based on stability until the maximum was found.
    \item \textbf{Max Angular Velocity (ANG-CAP)}: Measures the maximum absolute angular velocity under the same stability criteria and measure strategy as LIN-CAP.
    \item \textbf{Object Diversity (DIV-ROB)}: Tests the trained policy on objects with varying weights and textures to assess robustness against different dynamics. For chairs, three variants with different textures and weights are used. For tables, two different tables are tested. For racks, a rack and a standing lamp are used for comparison. Success is measured by the robot moving these objects at medium speed for 8 seconds. For every object, we run 10 trials, and the score is measure by the success rate. $v^{DIV-ROB} = \frac{1}{10 N} 
    \sum_1^N\mathbf{1} $(if success). We list the objects we used for this experiment in the appendix.
    \item \textbf{Initial Condition Sensitivity (IC-ROB)}: Evaluates the sensitivity of the policy to variations in the initial contact condition. Small random perturbations are introduced, and the policy's performance is assessed under these changes. In this experiment, we randomly sampled 3 initial contact poses close to the default contact pose. For each initial contact pose, we run 10 trials, and the score is also measure by the success rate. This leads to a total of 3*10=30 evaluations for a method on a single type of object. 
    \item \textbf{Single Trajectory Tracking (STT-CON)}: Measures the endpoint difference between the robot's planned rolling trajectory and the actual trajectory during single trajectory execution.
    \item \textbf{Multiple Trajectory Tracking (MTT-CON)}: Measures the endpoint difference for a sequence of trajectories generated by rolling out a series of planned trajectories. Compare to STT-CON, the metric measure also measure the motion transition property. The evaluation is the same as we used in Section:\ref{sec:exp_sim_setup}.
\end{itemize}

\begin{figure*}[t]
    \centering
   \includegraphics[width=.99\linewidth]{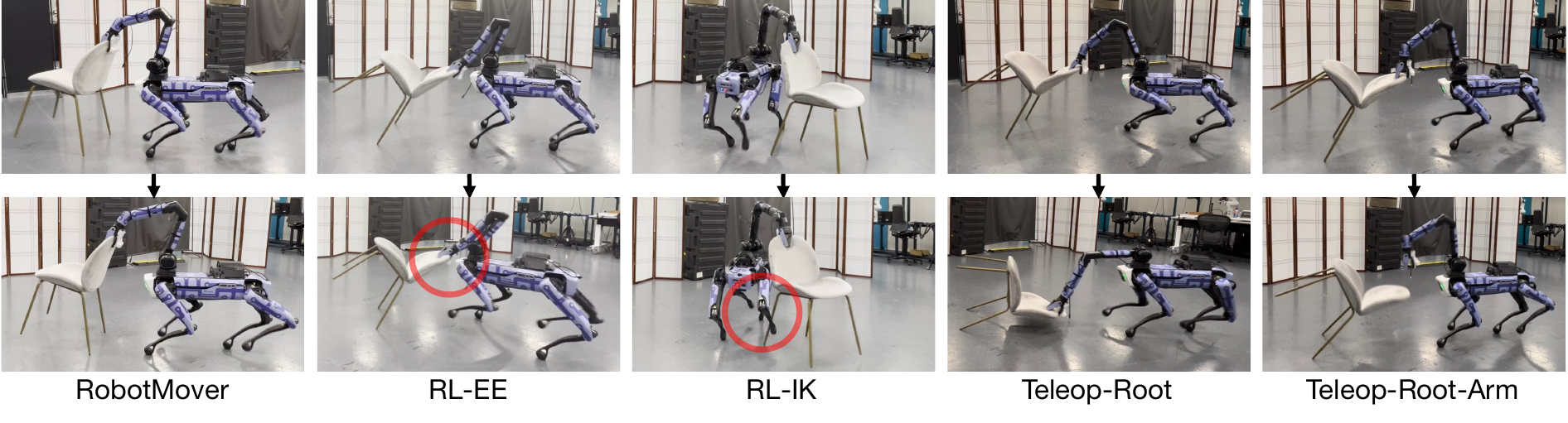}
   \caption{Robot moving a chair using different methods. \textbf{\method} shows stable chair moving. \textbf{RL-EE} and \textbf{RL-IK} result in robot self-collision and robot-object collision. \textbf{Teleop-Root} and \textbf{Teleop-Root-Arm} cannot stably move the chair, causing the chair to fall off. }
   \label{fig:hardware_visual}
    \vspace{-0.3cm}
   \end{figure*}

\subsection{Hardware Comparison Results}
\label{sec:hardware_result}

We summarize our quantitative hardware evaluation in Figure~\ref{fig:hardware_results}. Overall, \method \ outperforms all baselines over all metrics. For capability, \method \ achieves the highest object-moving velocities, with a maximum linear velocity of 0.75 m/s and an angular velocity of 0.5 rad/s, while other methods struggle to maintain stability at such speeds, often resulting in the object falling off. In terms of robustness, our method demonstrates lower sensitivity to variations in object properties (e.g., weight, friction, thickness) and initial contact poses. Such variations introduce differences in inertia and contact forces, which can negatively impact policy performance. However, the learned \method \ policy successfully handles to these dynamic changes, proving its potential for real-world application where the policy need to move diverse objects.
In our experiment, \method policy is the only policy that able to move a large table and a heavy rack with success rate over 50\%. For controllability, our result shows \method \ achieves the best object velocity tracking performance over all types of objects. Other methods suffer either large sim2real gap or unsuitable root arm coordination which lead to inaccurate velocity tracking.

A qualitative comparison between \method \ and learning-based methods is shown in Figure~\ref{fig:hardware_visual}. Compared to the learning-based baselines (\textbf{RL-EE} and \textbf{RL-IK}), \method \ leverages a interaction chain representation, which leads to better task performance and facilitates sim-to-real transfer.  Although \textbf{RL-EE} achieves good performance in simulation for chair and table manipulation, the learned policy's desired arm pose on hardware often results in self-collisions, causing damage to the robot and leading to objects slipping off. Similarly, \textbf{RL-IK} exhibits poor hardware performance, as the policy frequently causes the object to collide with the robot’s rear legs, significantly degrading overall task execution as well as raising safty concerns to robot.

When compared to teleoperation methods, \method \ also demonstrates clear advantages. Teleoperation heavily depends on the operator's skill level, leading to lower capability and robustness under our experimental settings. Specifically, the \method \ policy controls both the arm and root movements simultaneously while measuring the robot's state at 20~Hz. This allows it to responsively adjust its actions to compensate for disturbances or oscillations—an ability that is nearly impossible for a human operator to achieve with the same level of precision. Additionally, in terms of controllability, teleoperation strategies primarily rely on root-level commands for the robot, whereas \method’s \ policy operates in the object’s coordinate frame. This design enables \method \ to achieve significantly better tracking performance. Figure~\ref{fig:hardware_visual} illustrates failure cases of the teleoperation methods. In particular, \textbf{Teleop-Root} fails to execute the appropriate torque, causing the chair to fall off. Moreover, \textbf{Teleop-Root-Arm} overemphasizes arm pose and root velocity, failing to maintain consistent movement of the chair.

\subsection{Real World Applications}
\label{sec:applications}

\begin{figure*}[t]
    \centering
        \includegraphics[width=.9\linewidth]{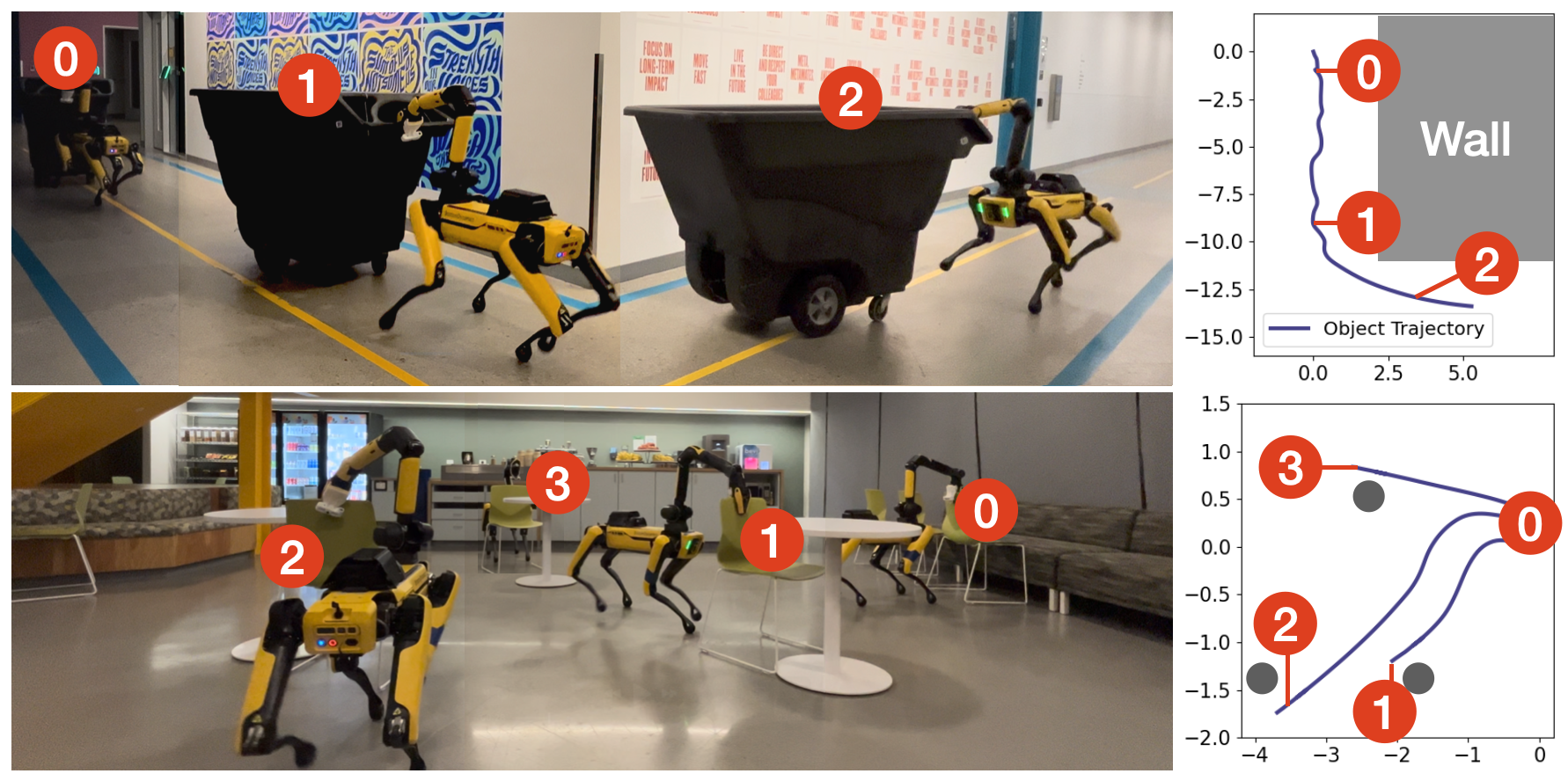}

       \caption{Illustration of two applications: Trash Cart Transportation (top) and Chair Rearrangement (bottom). A motion planner determines the object trajectory for each task (right), our trained policy controls the robot to move the object of interest (left).}
\label{fig:applications}
 \vspace{-0.6cm}
\end{figure*}

We demonstrate the capability of the learned policies to address complex real-world tasks by integrating them with high-level planners. Two applications are presented: \textit{Trash Cart Transportation} and \textit{Chair Rearrangement}. Trash cart transportation is a common task in office and residential environments where a loaded trash cart needs to be transported from one place to another. Chair rearrangement universally important task in various social and public settings, such as dining areas, meeting rooms, and event spaces. In our experiment, we integrate a learned object manipulation policy with a teleoperation-based high-level planner, we present interactive trash cart transportation. By integrating the policy with a heuristic object root velocity planner, we showcase automatic chair rearrangement. 
Snapshots illustrating these tasks are provided in Figure~\ref{fig:applications}.

In the \textit{Trash Cart Transportation} experiment, the robot's objective is to drag a large trash cart backward for approximately 10 meters along a hallway and then execute a 90-degree turn to transition to a perpendicular path. The trash cart, measuring approximately 1.8~m × 1.0~m × 1.2~m (L × W × H) and equipped with four wheels, is significantly larger than the robot itself. When manipulating the cart using the trained policy, the cart does not perfectly track the velocity command due to the asymmetrical weight distribution of the loaded cart and variations in contact status during the robot's stepping motion. 

To address this challenge, we employ an interactive control strategy. The user selects commands from a predefined set of temporary trajectories (a ``cookbook'') using keyboard inputs. These commands dynamically guide the robot, ensuring the trash cart follows the desired path. Notably, the policy used in this experiment is the same policy trained for moving chairs, which are relatively lighter than the trash cart.

The results show that the large trash cart is successfully transported along the desired path. The trained policy adapts to the cart's dynamics without fine-tuning. Notably, the robot’s root-arm coordination is evident during the object turning stage, where the robot follows a large arc while simultaneously adjusting the arm pose to guide the cart along a smaller arc. Our findings demonstrate that by combining the learned policy with an interactive control framework, the robot can reliably drag the trash cart along the specified route.

In the \textit{Chair Rearrangement} experiment, the robot has to rearrange three chairs in a dining space of approximately 5~m x 4~m (L × W). The chairs are initially in the same location but each one has a different goal positions and orientations (headings) relative to the global frame. The goal positions and orientations are predetermined. 

To achieve automatic chair rearrangement, we use a rule-based heuristic high-planner planner. This planner runs at 2Hz and commands target object velocities by comparing the object’s current global position and heading with the desired goal state. We describe the implementation of the heuristic planner in more detail in the Appendix. Spot’s onboard sensing system is utilized to determine the global positions of the robot and the chairs, while the end-effector position is used to approximate the objects' positions. For simplicity, we assume that there are no obstacles between the starting positions and the target locations. Once a chair reaches its goal position, the robot is teleoperated to grasp and move the next chair.

Our results show that the robot can automatically place the chair in different goal states. On average, the robot takes approximately 30 seconds to move a chair 5 meters away while adjusting for a heading difference of $\pi/2$ radians. The system achieves an average tracking error of 0.15 meters in position and 0.3 radians in orientation. In addition, the chair used in this experiment has a different shape compared to the one used during training. Specifically, the experimental chair has a thinner back (4 mm in the experiment versus 10 mm in simulation), resulting in reduced contact force between the robot's gripper and the chair. Lower contact force can potentially lead to reduced friction, increasing the likelihood of the chair slipping off the robot. However, the policy successfully adapts to this variation. We argue that this is due to the inclusion of friction randomization during training, enabling the policy to handle low-friction conditions effectively. These results highlight the system’s ability to perform precise and reliable chair rearrangement, underscoring the feasibility of automating such tasks in real-world environments.

\subsection{Two-Arm Demonstrations}
\label{sec:2-arms}

In demonstrations where a human uses both arms to manipulate an object, the resulting Interaction Chains cannot be directly used to train a single-arm robot, such as Spot, due to embodiment differences. To address this, we explore two approaches that enable learning from two-arm demonstrations: Chain Aggregation and Multi-Robot Collaboration. We evaluate both methods in simulation for the task of moving a coffee table using two-arm demonstrations.

\textbf{Chain Aggregation:}
In this approach, we aggregate the Interaction Chains from both arms into a single chain that can be used to train a single-arm robot. The aggregation is performed by taking the midpoint between corresponding nodes on the left and right arm chains: 
\begin{align} 
    \label{eqn:two_arms} 
    x^{i,\text{agg}}_t = \frac{1}{2}(x^{i,l}_t + x^{i,r}_t), 
\end{align} 
where $x^{i,\text{agg}}_t$, $x^{i,l}_t$, and $x^{i,r}_t$ denote the position of the $i$-th node on the aggregated, left-arm, and right-arm interaction chains, respectively, at time $t$. Since both original chains start at the object root and terminate at the human root, the aggregated chain preserves the same endpoints.
We find that this method performs well when the human demonstrates symmetric arm poses, achieving a tracking score of $0.66 \pm 0.10$. However, in the asymmetric case, where each arm performs a different role, the performance drops significantly to $0.35 \pm 0.07$. This suggests that naive averaging of asymmetric poses can lead to ambiguous or physically implausible interaction references, making it less suitable for learning robust policies.

\begin{figure}[t]
    \centering
    \includegraphics[width=.995\linewidth]{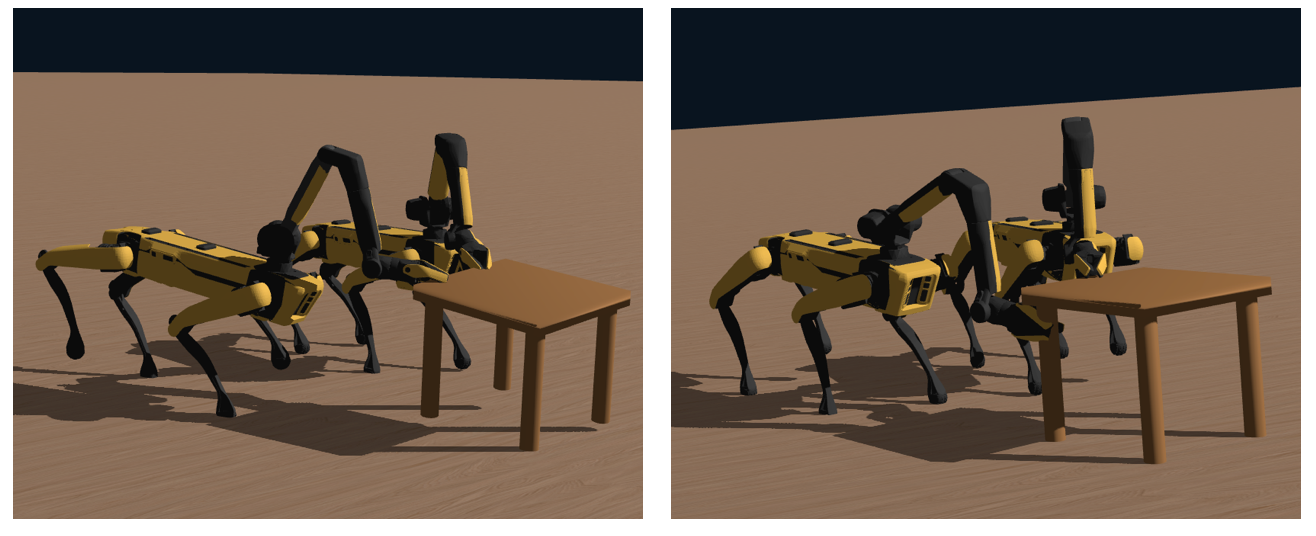}
    \caption{Two robots learning  to move a coffee table from two-arm demonstrations. Left: learned policy from symmetric arm poses. Right: learned policy from asymmetric arm poses.}
    \label{fig:two-robots}
    \vspace{-0.5cm}
\end{figure}

\textbf{Multi-Robot Collaboration:}
\added{As an alternative, we consider a collaborative setup using two single-arm robots, each assigned to imitate one arm from the human demonstration. A centralized controller is used to jointly control both robots, taking as input the combined proprioceptive states and outputting actions for both. The reward is computed as the sum of the two individual robot imitation rewards, with each robot assigned to either the left- or right-arm Interaction Chain. Fig.~\ref{fig:two-robots} provides a visualization of the learned policy, this collaborative method achieves a tracking score of $0.70 \pm 0.14$ and generalizes well across both symmetric and asymmetric demonstrations.These results indicate that the Interaction Chain representation can support multi-robot collaboration, effectively overcoming embodiment limitations in cases where a single robot cannot directly replicate the demonstrated behavior.
% \sehoon{we may report both symmetric and asymmetic cases and compare them against the chain aggregation. A small table can be helpful.} 
}

\section{Limitations and Future Work}
\label{sec:limitation}
Although our method achieves promising results, there are two limitations that we aim to address in future work.

First, our current framework places limited emphasis on the object-grasping process. In this work, we focus primarily on how the robot dynamically manipulates the object after establishing contact, rather than the grasp acquisition phase itself. In simulation, we initialize the robot’s pose so that the gripper is positioned near the object, allowing blind contact without active sensing. On hardware, the robot is manually positioned to achieve an initial grasp. Furthermore, to maintain high control frequency and because the robot’s onboard camera view is often obstructed after grasping, our control policy does not incorporate visual feedback. \added{This reliance on manual initialization limits the system’s autonomy in real-world applications.} In future work, we plan to develop a vision-based perception and control system that enables the robot to localize the object, navigate toward it, and establish a robust grasp autonomously, thereby providing initial contact conditions for the object-moving policy in a fully automatic manner.

Second, the generalization capabilities of the trained policies are currently constrained. Although the proposed imitation framework generalizes well across demonstrations for specific object type and target velocities, we train separate policies for different object categories. As a result, each policy is specialized to objects with similar shape and does not generalize across widely different object types. In future work, we aim to develop a unified policy that can handle a diverse range of objects and dynamically adapt its manipulation strategies based on the object’s physical characteristics. Toward this goal, we plan to explore more expressive policy architectures, such as diffusion models, and to incorporate richer object representations that capture variations in dynamics, contact properties, and shape. By doing so, we hope to extend the scalability and practical applicability of the RobotMover framework to more complex, real-world environments.

\section{Conclusion} 
\label{sec:conclusion}

We introduce \method, a framework that enables robots to manipulate large objects in the real world by training in simulation with human demonstrations. Critical to our approach is the Interaction Chain, a novel graphical representation that enables imitation of human-object interactions across different robot morphologies, facilitating the learning of reusable control policies. Our experimental evaluation in simulation demonstrates that policies trained with the Interaction Chain significantly outperform traditional robot learning methods in manipulating a variety of objects. Furthermore, our hardware experiments show that the proposed method produces more reliable object manipulation behaviors compared to both learning-based and teleoperation baselines. Finally, we integrate the learned policies with high-level planners, demonstrating the framework’s applicability to diverse real-world daily tasks.These results establish a strong foundation for scalable, real-world large object manipulation and open new avenues for future research on generalizable, cross-embodiment robot learning.

\bibliographystyle{IEEEtran}
\bibliography{references}

\begin{thebibliography}{10}
\providecommand{\url}[1]{#1}
\csname url@samestyle\endcsname
\providecommand{\newblock}{\relax}
\providecommand{\bibinfo}[2]{#2}
\providecommand{\BIBentrySTDinterwordspacing}{\spaceskip=0pt\relax}
\providecommand{\BIBentryALTinterwordstretchfactor}{4}
\providecommand{\BIBentryALTinterwordspacing}{\spaceskip=\fontdimen2\font plus
\BIBentryALTinterwordstretchfactor\fontdimen3\font minus \fontdimen4\font\relax}
\providecommand{\BIBforeignlanguage}[2]{{%
\expandafter\ifx\csname l@#1\endcsname\relax
\typeout{** WARNING: IEEEtran.bst: No hyphenation pattern has been}%
\typeout{** loaded for the language `#1'. Using the pattern for}%
\typeout{** the default language instead.}%
\else
\language=\csname l@#1\endcsname
\fi
#2}}
\providecommand{\BIBdecl}{\relax}
\BIBdecl

\bibitem{SayCan}
M.~Ahn, A.~Brohan, N.~Brown, Y.~Chebotar, O.~Cortes, B.~David, C.~Finn, C.~Fu, K.~Gopalakrishnan, K.~Hausman \emph{et~al.}, ``Do as i can, not as i say: Grounding language in robotic affordances,'' \emph{Conference on Robot Learning}, 2022.

\bibitem{chi2023diffusion}
C.~Chi, Z.~Xu, S.~Feng, E.~Cousineau, Y.~Du, B.~Burchfiel, R.~Tedrake, and S.~Song, ``Diffusion policy: Visuomotor policy learning via action diffusion,'' \emph{The International Journal of Robotics Research}, p. 02783649241273668, 2023.

\bibitem{black2024pi_0}
K.~Black, N.~Brown, D.~Driess, A.~Esmail, M.~Equi, C.~Finn, N.~Fusai, L.~Groom, K.~Hausman, B.~Ichter \emph{et~al.}, ``$pi\_0 $: A vision-language-action flow model for general robot control,'' \emph{arXiv preprint arXiv:2410.24164}, 2024.

\bibitem{team2024octo}
O.~M. Team, D.~Ghosh, H.~Walke, K.~Pertsch, K.~Black, O.~Mees, S.~Dasari, J.~Hejna, T.~Kreiman, C.~Xu \emph{et~al.}, ``Octo: An open-source generalist robot policy,'' \emph{arXiv preprint arXiv:2405.12213}, 2024.

\bibitem{MobileAloha}
Z.~Fu, T.~Z. Zhao, and C.~Finn, ``Mobile aloha: Learning bimanual mobile manipulation with low-cost whole-body teleoperation,'' \emph{{Conference on Robot Learning (CoRL)}}, 2024.

\bibitem{lu2024mobile}
C.~Lu, X.~Cheng, J.~Li, S.~Yang, M.~Ji, C.~Yuan, G.~Yang, S.~Yi, and X.~Wang, ``Mobile-television: Predictive motion priors for humanoid whole-body control,'' \emph{IEEE International Conference on Robotics and Automation (ICRA)}, 2025.

\bibitem{TidyBot}
J.~Wu, R.~Antonova, A.~Kan, M.~Lepert, A.~Zeng, S.~Song, J.~Bohg, S.~Rusinkiewicz, and T.~Funkhouser, ``Tidybot: Personalized robot assistance with large language models,'' \emph{Autonomous Robots}, vol.~47, no.~8, pp. 1087--1102, 2023.

\bibitem{Naoki_ASC}
N.~Yokoyama, A.~Clegg, J.~Truong, E.~Undersander, T.-Y. Yang, S.~Arnaud, S.~Ha, D.~Batra, and A.~Rai, ``Asc: Adaptive skill coordination for robotic mobile manipulation,'' \emph{IEEE Robotics and Automation Letters}, vol.~9, no.~1, pp. 779--786, 2023.

\bibitem{Behavior}
S.~Srivastava, C.~Li, M.~Lingelbach, R.~Mart{\'\i}n-Mart{\'\i}n, F.~Xia, K.~E. Vainio, Z.~Lian, C.~Gokmen, S.~Buch, K.~Liu \emph{et~al.}, ``Behavior: Benchmark for everyday household activities in virtual, interactive, and ecological environments,'' in \emph{Conference on robot learning}.\hskip 1em plus 0.5em minus 0.4em\relax PMLR, 2022, pp. 477--490.

\bibitem{Habitat}
M.~Savva, A.~Kadian, O.~Maksymets, Y.~Zhao, E.~Wijmans, B.~Jain, J.~Straub, J.~Liu, V.~Koltun, J.~Malik \emph{et~al.}, ``Habitat: A platform for embodied ai research,'' in \emph{Proceedings of the IEEE/CVF international conference on computer vision}, 2019, pp. 9339--9347.

\bibitem{mittal2022articulated}
M.~Mittal, D.~Hoeller, F.~Farshidian, M.~Hutter, and A.~Garg, ``Articulated object interaction in unknown scenes with whole-body mobile manipulation,'' in \emph{2022 IEEE/RSJ international conference on intelligent robots and systems (IROS)}.\hskip 1em plus 0.5em minus 0.4em\relax IEEE, 2022, pp. 1647--1654.

\bibitem{DeepWholeBody}
Z.~Fu, X.~Cheng, and D.~Pathak, ``Deep whole-body control: learning a unified policy for manipulation and locomotion,'' in \emph{Conference on Robot Learning}.\hskip 1em plus 0.5em minus 0.4em\relax PMLR, 2023, pp. 138--149.

\bibitem{RT1}
A.~Brohan, N.~Brown, J.~Carbajal, Y.~Chebotar, J.~Dabis, C.~Finn, K.~Gopalakrishnan, K.~Hausman, A.~Herzog, J.~Hsu \emph{et~al.}, ``Rt-1: Robotics transformer for real-world control at scale,'' \emph{arXiv preprint arXiv:2212.06817}, 2022.

\bibitem{krotkov2018darpa}
E.~Krotkov, D.~Hackett, L.~Jackel, M.~Perschbacher, J.~Pippine, J.~Strauss, G.~Pratt, and C.~Orlowski, ``The darpa robotics challenge finals: Results and perspectives,'' \emph{The DARPA robotics challenge finals: Humanoid robots to the rescue}, pp. 1--26, 2018.

\bibitem{wyrobek2008towards}
K.~A. Wyrobek, E.~H. Berger, H.~M. Van~der Loos, and J.~K. Salisbury, ``Towards a personal robotics development platform: Rationale and design of an intrinsically safe personal robot,'' in \emph{2008 IEEE International Conference on Robotics and Automation}.\hskip 1em plus 0.5em minus 0.4em\relax IEEE, 2008, pp. 2165--2170.

\bibitem{khatib1996force}
O.~Khatib, K.~Yokoi, K.~Chang, D.~Ruspini, R.~Holmberg, A.~Casal, and A.~Baader, ``Force strategies for cooperative tasks in multiple mobile manipulation systems,'' in \emph{Robotics Research: The Seventh International Symposium}.\hskip 1em plus 0.5em minus 0.4em\relax Springer, 1996, pp. 333--342.

\bibitem{garrett2021integrated}
C.~R. Garrett, R.~Chitnis, R.~Holladay, B.~Kim, T.~Silver, L.~P. Kaelbling, and T.~Lozano-P{\'e}rez, ``Integrated task and motion planning,'' \emph{Annual review of control, robotics, and autonomous systems}, vol.~4, no.~1, pp. 265--293, 2021.

\bibitem{sun2022fully}
C.~Sun, J.~Orbik, C.~M. Devin, B.~H. Yang, A.~Gupta, G.~Berseth, and S.~Levine, ``Fully autonomous real-world reinforcement learning with applications to mobile manipulation,'' in \emph{Conference on Robot Learning}.\hskip 1em plus 0.5em minus 0.4em\relax PMLR, 2022, pp. 308--319.

\bibitem{Homerobot}
S.~Yenamandra, A.~Ramachandran, K.~Yadav, A.~Wang, M.~Khanna, T.~Gervet, T.-Y. Yang, V.~Jain, A.~W. Clegg, J.~Turner \emph{et~al.}, ``Homerobot: Open-vocabulary mobile manipulation,'' \emph{arXiv preprint arXiv:2306.11565}, 2023.

\bibitem{ha2024umi}
H.~Ha, Y.~Gao, Z.~Fu, J.~Tan, and S.~Song, ``Umi on legs: Making manipulation policies mobile with manipulation-centric whole-body controllers,'' \emph{{Conference on Robot Learning (CoRL)}}, 2024.

\bibitem{xiong2024adaptive}
H.~Xiong, R.~Mendonca, K.~Shaw, and D.~Pathak, ``Adaptive mobile manipulation for articulated objects in the open world,'' \emph{arXiv preprint arXiv:2401.14403}, 2024.

\bibitem{shafiullah2023bringing}
N.~M.~M. Shafiullah, A.~Rai, H.~Etukuru, Y.~Liu, I.~Misra, S.~Chintala, and L.~Pinto, ``On bringing robots home,'' \emph{arXiv preprint arXiv:2311.16098}, 2023.

\bibitem{yang2023moma}
T.~Yang, Y.~Jing, H.~Wu, J.~Xu, K.~Sima, G.~Chen, Q.~Sima, and T.~Kong, ``Moma-force: Visual-force imitation for real-world mobile manipulation,'' in \emph{2023 IEEE/RSJ International Conference on Intelligent Robots and Systems (IROS)}.\hskip 1em plus 0.5em minus 0.4em\relax IEEE, 2023, pp. 6847--6852.

\bibitem{mendonca2024continuously}
R.~Mendonca, E.~Panov, B.~Bucher, J.~Wang, and D.~Pathak, ``Continuously improving mobile manipulation with autonomous real-world rl,'' \emph{{Conference on Robot Learning (CoRL)}}, 2024.

\bibitem{ravan2024combining}
Y.~Ravan, Z.~Yang, T.~Chen, T.~Lozano-P{\'e}rez, and L.~P. Kaelbling, ``Combining planning and diffusion for mobility with unknown dynamics,'' \emph{2023 IEEE International Conference on Robotics and Automation (ICRA)}, 2025.

\bibitem{jang2022bc}
E.~Jang, A.~Irpan, M.~Khansari, D.~Kappler, F.~Ebert, C.~Lynch, S.~Levine, and C.~Finn, ``Bc-z: Zero-shot task generalization with robotic imitation learning,'' in \emph{Conference on Robot Learning}.\hskip 1em plus 0.5em minus 0.4em\relax PMLR, 2022, pp. 991--1002.

\bibitem{li2023crossloco}
T.~Li, H.~Jung, M.~Gombolay, Y.~K. Cho, and S.~Ha, ``Crossloco: Human motion driven control of legged robots via guided unsupervised reinforcement learning,'' \emph{International Conference on Learning Representations}, 2024.

\bibitem{shafiullah2022behavior}
N.~M. Shafiullah, Z.~Cui, A.~A. Altanzaya, and L.~Pinto, ``Behavior transformers: Cloning $ k $ modes with one stone,'' \emph{Advances in neural information processing systems}, vol.~35, pp. 22\,955--22\,968, 2022.

\bibitem{li2023ace}
T.~Li, J.~Won, A.~Clegg, J.~Kim, A.~Rai, and S.~Ha, ``Ace: Adversarial correspondence embedding for cross morphology motion retargeting from human to nonhuman characters,'' in \emph{SIGGRAPH Asia 2023 Conference Papers}, 2023, pp. 1--11.

\bibitem{smith2023learning}
L.~Smith, J.~C. Kew, T.~Li, L.~Luu, X.~B. Peng, S.~Ha, J.~Tan, and S.~Levine, ``Learning and adapting agile locomotion skills by transferring experience,'' \emph{Robotics: Science and Systems}, 2023.

\bibitem{vuong2023open}
Q.~Vuong, S.~Levine, H.~R. Walke, K.~Pertsch, A.~Singh, R.~Doshi, C.~Xu, J.~Luo, L.~Tan, D.~Shah \emph{et~al.}, ``Open x-embodiment: Robotic learning datasets and rt-x models,'' in \emph{Towards Generalist Robots: Learning Paradigms for Scalable Skill Acquisition@ CoRL2023}, 2023.

\bibitem{chi2024universal}
C.~Chi, Z.~Xu, C.~Pan, E.~Cousineau, B.~Burchfiel, S.~Feng, R.~Tedrake, and S.~Song, ``Universal manipulation interface: In-the-wild robot teaching without in-the-wild robots,'' \emph{Robotics: Science and Systems}, 2024.

\bibitem{li2019using}
T.~Li, H.~Geyer, C.~G. Atkeson, and A.~Rai, ``Using deep reinforcement learning to learn high-level policies on the atrias biped,'' in \emph{2019 International Conference on Robotics and Automation (ICRA)}.\hskip 1em plus 0.5em minus 0.4em\relax IEEE, 2019, pp. 263--269.

\bibitem{RoboImitationPeng20}
X.~B. Peng, E.~Coumans, T.~Zhang, T.-W.~E. Lee, J.~Tan, and S.~Levine, ``Learning agile robotic locomotion skills by imitating animals,'' in \emph{Robotics: Science and Systems}, 07 2020.

\bibitem{peng2018deepmimic}
X.~B. Peng, P.~Abbeel, S.~Levine, and M.~Van~de Panne, ``Deepmimic: Example-guided deep reinforcement learning of physics-based character skills,'' \emph{ACM Transactions On Graphics (TOG)}, vol.~37, no.~4, pp. 1--14, 2018.

\bibitem{peng2021amp}
X.~B. Peng, Z.~Ma, P.~Abbeel, S.~Levine, and A.~Kanazawa, ``Amp: Adversarial motion priors for stylized physics-based character control,'' \emph{ACM Transactions on Graphics (ToG)}, vol.~40, no.~4, pp. 1--20, 2021.

\bibitem{peng2022ase}
X.~B. Peng, Y.~Guo, L.~Halper, S.~Levine, and S.~Fidler, ``Ase: Large-scale reusable adversarial skill embeddings for physically simulated characters,'' \emph{ACM Transactions On Graphics (TOG)}, vol.~41, no.~4, pp. 1--17, 2022.

\bibitem{InteractionGraph}
Y.~Zhang, D.~Gopinath, Y.~Ye, J.~Hodgins, G.~Turk, and J.~Won, ``Simulation and retargeting of complex multi-character interactions,'' in \emph{ACM SIGGRAPH 2023 Conference Proceedings}, 2023, pp. 1--11.

\bibitem{ho2010spatial}
E.~S. Ho, T.~Komura, and C.-L. Tai, ``Spatial relationship preserving character motion adaptation,'' in \emph{ACM SIGGRAPH 2010 papers}, 2010, pp. 1--8.

\bibitem{rudin2022learning}
N.~Rudin, D.~Hoeller, P.~Reist, and M.~Hutter, ``Learning to walk in minutes using massively parallel deep reinforcement learning,'' in \emph{Conference on Robot Learning}.\hskip 1em plus 0.5em minus 0.4em\relax PMLR, 2022, pp. 91--100.

\bibitem{PPO}
J.~Schulman, F.~Wolski, P.~Dhariwal, A.~Radford, and O.~Klimov, ``Proximal policy optimization algorithms,'' \emph{arXiv preprint arXiv:1707.06347}, 2017.

\bibitem{OMOMO}
J.~Li, J.~Wu, and C.~K. Liu, ``Object motion guided human motion synthesis,'' \emph{ACM Transactions on Graphics (TOG)}, vol.~42, no.~6, pp. 1--11, 2023.

\bibitem{Genesis}
\BIBentryALTinterwordspacing
G.~Authors, ``Genesis: A universal and generative physics engine for robotics and beyond,'' December 2024. [Online]. Available: \url{https://github.com/Genesis-Embodied-AI/Genesis}
\BIBentrySTDinterwordspacing

\bibitem{ELUs}
D.-A. Clevert, ``Fast and accurate deep network learning by exponential linear units (elus),'' \emph{arXiv preprint arXiv:1511.07289}, 2015.

\end{thebibliography}

\newpage
\section{Appendix}
\subsection{Dataset Details}
We selected human-object interaction demonstrations from the OMOMO dataset~\citep{OMOMO}, a motion capture-based dataset featuring high-quality and diverse human and object movements. Although OMOMO contains demonstrations for more than ten objects, some, such as monitors, are not relevant to our task and were excluded. Additionally, objects that require significant two-arm manipulation, such as large boxes, were omitted to align with our study's focus. Below, we summarize the number of demonstration trajectories used for training and their corresponding sources in the OMOMO dataset.

\vspace{0.2cm}

\begin{tabular}{c | c | c }
\toprule
\rowcolor[HTML]{ededed}
Object Type& Num  &  Demo Source\\
\midrule
Chair & 50 &  ['woodchair', 'whitechair'] \\
\midrule
Table &  30 &  ['largetable', 'smalltable']  \\
\midrule
StandingStick & 40 &  ['floorlamp', 'clothesstand']\\
\bottomrule
\end{tabular}

\vspace{0.2cm}

\subsection{Diverse Objects Experiment}

\begin{figure}[H]
    \begin{minipage}[b]{.495\textwidth}
        \centering
        \includegraphics[width=.995\linewidth]{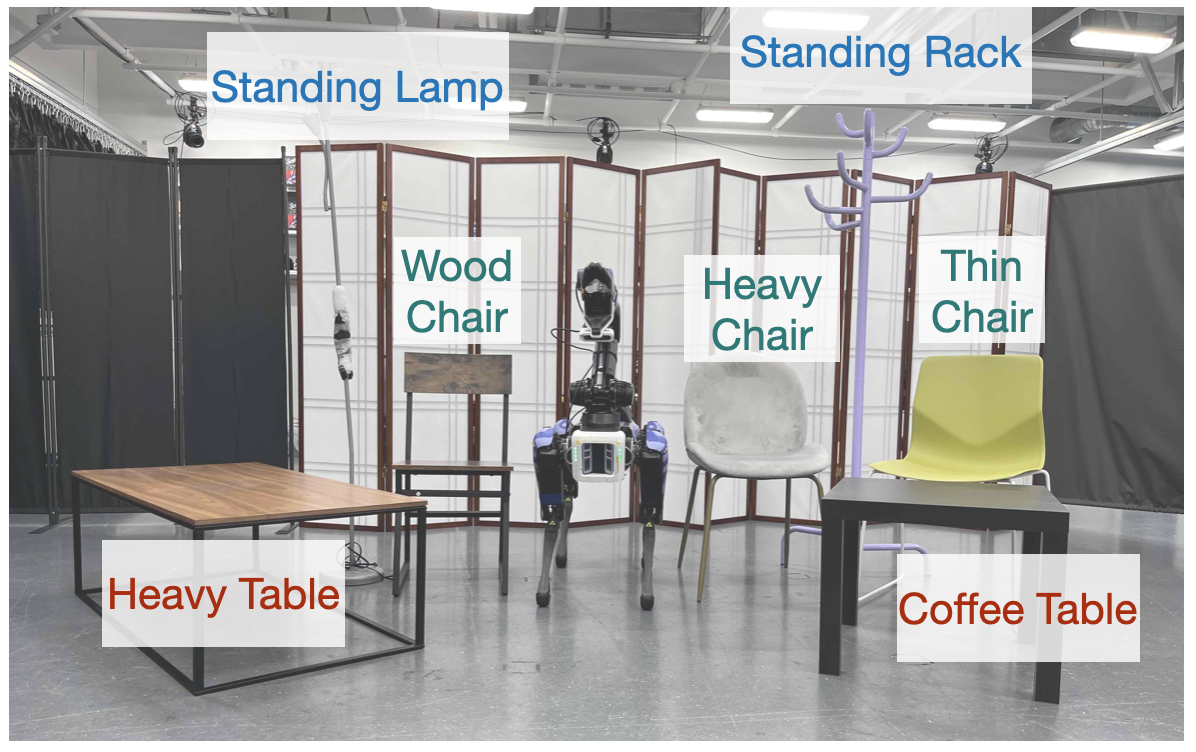}
    \end{minipage}

    \vspace{.25cm}
    \begin{minipage}[b]{.495\textwidth}
        \begin{tabular}{c | c | c }
            
            \toprule
            \rowcolor[HTML]{ededed}
            Name&  Size(cm) & Note\\
            \midrule
            Wood Chair & 49*44*80 &  Medium Weight, Low Friction \\
            \midrule
            Heavy Chair &  55*52*80 &  Heavy, Medium Friction  \\
            \midrule
            Thin Chair &  57*55*80 &  Medium Weight, Low Friction  \\
            \midrule
            Coffee Table &  55*55*45 &  Medium Weight, Medium Friction  \\
            \midrule
            Heavy Table &   57*144*42 &  Heavy, Low Friction  \\
            \midrule
            Standing Lamp &  20*20*175 &  Light Weight, Low Friction  \\
            \midrule
            Standing Rack & 60*60*175 & Heavy, Low Friction \\
            \bottomrule
        \end{tabular}
    \end{minipage}
\caption{The collection of objects used in the diverse object experiments.}
 \label{fig:diverse}
\end{figure}

In Section~\ref{sec:hardware_setup}, we introduce Object Diveristy(DIV-ROB) metrics to evaluate the robustness of the result policies by moving objects with different textures and weights. We summarize the objects for our diverse object test in Figure~\ref{fig:diverse}. 

\subsection{Heuristic Planner for Chair Rearrangement}
\label{app:heuristic_planner}
In the Chair Rearrangement experiment (Section~\ref{sec:applications}), we use a heuristic planner for object target velocity planning. The planner first moves the object to the target location and then aligns its heading. The detailed planning policy is design as follows:

\begin{algorithm}[htbp]
\caption{Heuristic Planner}
\label{alg:heurisitc_planner}
    \begin{algorithmic}[1]{
        \footnotesize
        \Require Target object position $\bar{p}^{xy}$ and heading $\bar{p}^{head}$, position threshold $\bar{\delta}^{xy}$, heading threshold $\bar{\delta}^{yaw}$, robot policy $\pi$
        % \State Initialize: object position and heading, robot policy $\pi$
        \item[]
            
        \State {{\textsc{// Target Position Reaching }}}
        % \If{ is empty}
        \Repeat
            \State Measure the object position $p^{xy}$ and the delta heading towards the target position $\hat{p}^{head}$.
            \State Target object velocity $\bar{v^{o}}=[0.4,0.0, min(0.4,\hat{p}^{head})] $.
            \State Run robot policy $\pi$.
            
        \Until{$||\bar{p}^{xy} - p^{xy}|| < \bar{\delta}^{xy}$}
        % \EndIf
        \item[]
            \State {{\textsc{// Target Heading Aligning}}}
      	\Repeat
             \State Measure heading difference $\delta^{head} = \bar{p}^{head} - p^{head}$.
            \State Target object velocity $\bar{v^{o}}=[0.0,0.0, min(0.4,\delta^{head}]$.
            \State Run robot policy $\pi$.
      	\Until{$||\delta^{head}|| < \bar{\delta}^{head}$}
            }
  \end{algorithmic}
\end{algorithm}

We found that this simple heuristic planner works well in our setting. However, for more complex tasks, such as those involving obstacle avoidance, a more sophisticated planner would be required.

\end{document}